\documentclass[twocolumn]{autart}    

\usepackage[utf8]{inputenc}
\usepackage{float,color,cite}
\usepackage{dsfont}
\usepackage{bm}
\usepackage{stmaryrd}
\usepackage{booktabs,tabularx, hhline,multicol}
\usepackage{amsmath}
\usepackage{amssymb,amsfonts}
\usepackage{textcomp, gensymb}
\usepackage{array}
\usepackage[ruled,vlined,linesnumbered]{algorithm2e}
\usepackage{algorithmic}
\usepackage{hyperref,xcolor}
\SetKwInOut{Parameter}{Parameters}
\usepackage{graphicx}
\usepackage{booktabs}
\usepackage{diagbox}
\usepackage{autobreak}
\usepackage{caption}
\usepackage{subcaption}
\usepackage{comment}
\usepackage[english]{babel}
\newtheorem{theorem}{Theorem}

\newtheorem{remark}{Remark}

\newcommand{\enumbracket}[1]{{\llbracket #1 \rrbracket}}

\newcommand{\Rmnum}[1]{\uppercase\expandafter{\romannumeral #1}}
\newcommand{\modelname}{PredVAR}
\newcommand{\rrname}{RRVAR}

\makeatletter
\newcommand*\bigcdot{\mathpalette\bigcdot@{.5}}
\newcommand*\bigcdot@[2]{\mathbin{\vcenter{\hbox{\scalebox{#2}{$\m@th#1\bullet$}}}}}
\makeatother

\begin{document}

\begin{frontmatter}
\title{Probabilistic Reduced-Dimensional Vector Autoregressive Modeling  with Oblique Projections \thanksref{footnoteinfo}}

\thanks[footnoteinfo]{The work described in this paper was partially supported by a grant from
a General Research Fund by the Research Grants Council (RGC) of Hong Kong SAR,
China (Project No. 11303421), a Collaborative Research Fund by RGC of Hong Kong (Project No. C1143-20G),
a grant from the Natural Science Foundation of China (U20A20189),
a grant from ITF - Guangdong-Hong Kong Technology Cooperation Funding Scheme (Project Ref. No. GHP/145/20),
a Math and Application Project (2021YFA1003504) under the National Key
R\&D Program, a Shenzhen-Hong Kong-Macau Science and Technology Project
Category C (9240086), and an InnoHK initiative of The Government of the
HKSAR for the Laboratory for AI-Powered Financial Technologies. Part of the material in this paper was presented at the 62nd IEEE Conference
on Decision and Control, December 13–15, 2023, Singapore~\cite{mo:yu:qin2023CDC}. \emph{(Corresponding author: S.~J.~Qin)}
        }

\author[CityU]{Yanfang~Mo}\ead{yanfang.mo@cityu.edu.hk}, and
\author[LU]{S.~Joe~Qin}\ead{joeqin@LN.edu.hk}

\address[CityU]{Hong Kong Institute for Data Science, City University of Hong Kong, Hong Kong}
\address[LU]{The Institute of Data Science, Lingnan University, Hong Kong}


\begin{abstract}
In this paper, we propose a probabilistic reduced-dimensional vector autoregressive (PredVAR) model to extract low-dimensional dynamics from high-dimensional noisy data. The model utilizes an oblique projection to partition the measurement space into a subspace that accommodates the reduced-dimensional dynamics and a complementary static subspace. An optimal oblique decomposition is derived for the best predictability regarding prediction error covariance. Building on this, we develop an iterative PredVAR algorithm using maximum likelihood and the expectation-maximization (EM) framework. This algorithm alternately updates the estimates of the latent dynamics and optimal oblique projection, yielding dynamic latent variables with rank-ordered predictability and an explicit latent VAR model that is consistent with the outer projection model. The superior performance and efficiency of the proposed approach are demonstrated using data sets from a synthesized Lorenz system and an industrial process from Eastman Chemical.
\end{abstract}

\end{frontmatter}

\section{Introduction}

With the popular deployment of the industrial internet of things, high-dimensional data with low-dimensional dynamics~\cite{sznaier2020control}  pose new challenges to traditional time series analysis methods that assume fully excited full-dimensional dynamics.
In many real industrial operations,  high dimensional time series often do not exhibit full-dimensional dynamics~\cite{pena1987identifying}. In addition, autonomous systems deploy redundant sensors for safety and fault tolerance, resulting in high collinearity in the collected time-series data. Moreover, computationally intensive applications like digital twins produce high dimensional data to capture important dynamics for effective decision-making~\cite{fries2022lasdi}. These popular application scenarios call for parsimonious modeling to extract reduced-dimensional dynamics in high dimensional data~\cite{qin2020bridging,reinsel:velu:1998}.

Classic reduced-dimensional analytic tools, such as principal component analysis (PCA), partial least squares (PLS) \cite{geladi:kowalski:1986, qin:liu:tang:2023SDPLS}, and canonical correlation analysis (CCA), have been successful in many applications. While they perform dimension reduction, their statistical inference requires serial independence in the data~\cite{qin2020bridging}.

Many efforts in the statistical field have raised the issue of reduced dimensional dynamics in multivariate time series data. An early account of this problem is given in \cite{box:tiao:1977}, which does the dynamics modeling and dimensional reduction in two steps and thus is not optimal. Subsequently,
 dynamic factor models (DFMs) are developed, which rely on time-related statistics to estimate parsimonious model parameters, including~\cite{pena1987identifying,lam2012factor,pena2019forecasting,gao:tsay:2021high-Dim}. However, these DFMs do not enforce a parsimonious latent dynamic model. The linear Gaussian state-space model in~\cite{wen2012data} and the autoregressive DLV model in~\cite{zhou2016autoregressive} extract the dynamic and static characteristics simultaneously. Nevertheless, they ignore the structured signal-noise relationship and are not concerned with the dimension reduction in noise.

In process data analytics,
 a dynamic latent variable (DLV) model was developed in~\cite{li2014new} to extract DLVs first and characterize the dynamic relations in the DLVs and the static cross-correlations in residuals afterward. Assuming the latent dynamics is integrating or being `slow', slow feature analysis has been developed \cite{Zhao:huang:2018SFA,Shang:etal:2018Recursive-SFA,fan:etal:huang:2019SFA} to induce dimension reduction focusing on slow dynamics only.
 To develop full latent dynamic features, dynamic-inner PCA (DiPCA)~\cite{dong2018novel} and dynamic-inner CCA (DiCCA)~\cite{dong2018dynamic,dong:liu:qin:2020DICCA-SVD} are proposed to produce rank-ordered DLVs to maximize the prediction power in terms of covariance and canonical correlation, respectively. Furthermore, Qin developed a latent vector autoregressive modeling algorithm with a CCA objective (LaVAR-CCA)~\cite{qin2021latent,Qin:2022LaVAR_AIChEJ}, whose state-space generalization is developed  in~\cite{yu:qin:2022LaSS}.
Casting the LaVAR model in a probabilistic setting, \cite{ZhuQQ:etal:2022PPFA} uses the Kalman filter formulation to solve for the dynamic and static models via maximum likelihood, but fails short of solving the dynamic model parameters explicitly and resorts to genetic algorithms for the solution. A key to signal reconstruction and prediction lies in separating the serially dependent signals from serially independent noise that may  have energy or variance comparable to that of the signals~\cite{qin2020bridging}.

The prevalence of low-dimensional dynamics in high-dimensional uncertainties requires a statistical framework for simultaneous  dimension reduction and latent dynamics extraction. In this paper, we propose a probabilistic reduced-dimensional vector autoregressive (PredVAR) model to extract low-dimensional dynamics from high-dimensional noisy data.
Our probabilistic model partitions the measurement space into a  low-dimensional DLV subspace and a static noise subspace, which are generally oblique. An oblique projection is adopted to delineate the structural signal-noise relationship, extending the widely used orthogonal projection~\cite{lou2022novel}. The oblique projection for the dynamic-static decomposition in our model is found simultaneously with an explicit low-dimensional latent dynamic model.  Accordingly, our model consists of two interrelated components, namely, the optimal oblique projection and the latent dynamics. An expectation-maximization (EM) procedure is used to estimate model parameters~\cite{dempster1977maximum,yu2022generalized} that satisfy the maximum likelihood optimal conditions.

The main contributions of this work are as follows.

$1)$ A probabilistic reduced-dimensional vector regressive (\modelname) model is proposed with general oblique projections. Our study of using VAR to capture latent dynamics should initiate exploring more dynamic models.

$2)$ A particular oblique projection is derived for the optimal dynamic-static decomposition, which achieves the best predictability regarding prediction error covariance and has an intriguing geometric interpretation. Also, ordered DLVs are obtained based on predictability.

$3)$ A \modelname~model identification algorithm is developed based on the optimal dynamic-static decomposition, the interplay between latent dynamics and projection identifications, and the EM method. The noise covariance matrices are estimated as a byproduct of the EM method and the analytical comparison with LaVAR-CCA and DiCCA naturally follows.

$4)$ Extensive simulations are presented to illustrate our approach compared to conceivable benchmarks over the synthesized Lorenz data and real Eastman process data.

The rest of the paper is organized as follows. The \modelname~model is formulated in Section~\ref{sec:model}. An efficient \modelname~model identification algorithm is developed in Section~\ref{sec:alg}, based on the optimal dynamic-static decomposition introduced in Section~\ref{sec:opt_dyn_static_decomp}. Additional analysis is presented in Section~\ref{sec:add_analysis}, including the distributions of parameter estimates and the selection of model sizes. Section~\ref{sec:simulation} presents  simulation comparison and studies. Finally, this paper is concluded in Section~\ref{sec:conclusion}. Table~\ref{tab: notation} summarizes the major notation.

\begin{table}[t]
\centering
\caption{Notation.}
\vspace{.5\baselineskip}
  \label{tab: notation}
  \begin{tabular}{ll}
  \hline\hline
  $\enumbracket{n}$ & $\{1,2,\ldots, n\}$ for a natural number $n$\\
  $s$ & order of the vector autoregressive (VAR) dynamics\\
 $p$ & dimension of the measurement space\\
 $\ell$ & dynamic latent dimension (DLD)\\
 $\bm y_k$ & measurement vector in~$\Re^p$\\
 $\bm v_k$ & Dynamic latent variables (DLVs) in~$\Re^\ell $\\
 $\bm{\hat v}_k$ & one-step ahead DLVs prediction in~$\Re^\ell $\\
  $\bm \varepsilon_k$ & DLV innovations vector in~$\Re^{\ell }$, $\bm \varepsilon_k \sim \mathcal{N}(\bm 0, \mathbf{\Sigma}_{\bm \varepsilon})$\\
 $\bm{\bar \varepsilon}_k$ & outer model noise in~$\Re^{p-\ell }$, $\bm{\bar \varepsilon}_k \sim \mathcal{N}(\bm 0, \mathbf{\Sigma}_{\bm{\bar \varepsilon}})$\\
 $\bm e_k$ & innovations vector in~$\Re^{p}$,~$\bm e_k = \mathbf{P}\bm \varepsilon_k +\mathbf{\bar{P}}\bm{\bar \varepsilon}_k$\\
 $\mathbf{I}/ \bm 0$ & identity/zero matrix of a compatible dimension\\
 $\mathbf{B}_j$ &  VAR  coefficient matrix in~$\Re^{\ell \times \ell }$, $j\in \enumbracket{s}$\\
 $\mathbf{P}$ & DLV loadings matrix in~$\Re^{p\times \ell }$\\ 
 $\mathbf{\bar{P}}$ & static loadings matrix in~$\Re^{p\times {(p-\ell )}}$\\ 
 $\mathbf{R}$ & DLV weight matrix in~$\Re^{p\times \ell }$\\
 $\mathbf{\bar{R}}$ & static weight matrix in~$\Re^{p\times {(p-\ell )}}$\\ 
 $\mathbf{\Sigma}_{\bm{*}}$ & covariance matrix of a random vector~$\bm{*}$\\
 \hline \hline
  \end{tabular}
\end{table}

\section{Model Formulation}\label{sec:model}
\subsection{The PredVAR Model}

Consider a time series~$\{\bm{y}_k \in \Re^p\}_{k=1}^{N+s}$ with $E(\bm y_k) = \bm 0$ for all~$k\in \enumbracket{N+s}$. Qin \cite{Qin:2022LaVAR_AIChEJ} defines that the time series is serially correlated or dynamic if, for at least one $j > 0$, $${E} \left \{ \bm{y}_k \bm{y}_{k-j}^\intercal \right \} \neq \mathbf{0}.$$ Otherwise, it is serially uncorrelated. Further, as per Qin \cite{Qin:2022LaVAR_AIChEJ}, $\{\bm{y}_k\}$ is
 a reduced-dimensional dynamic (RDD) series if it is serially correlated but $\{\mathbf{a}^\intercal\boldsymbol{y}_k\}$ is serially uncorrelated for some $\mathbf{a \neq 0} \in \Re^p$. If no $\mathbf{a \neq 0}$ exists to render $\{\mathbf{a}^\intercal\bm{y}_k\}$ serially uncorrelated, $\{\bm{y}_k\}$ is full-dimensional dynamic (FDD). Qin \cite{Qin:2022LaVAR_AIChEJ} develops a latent VAR model to estimate the  RDD dynamics of the data series. However, no statistical analysis is given since the latent VAR model does not give the distributions of noise terms.

\begin{figure}[t]
    \centering
\includegraphics[width=0.9\columnwidth]{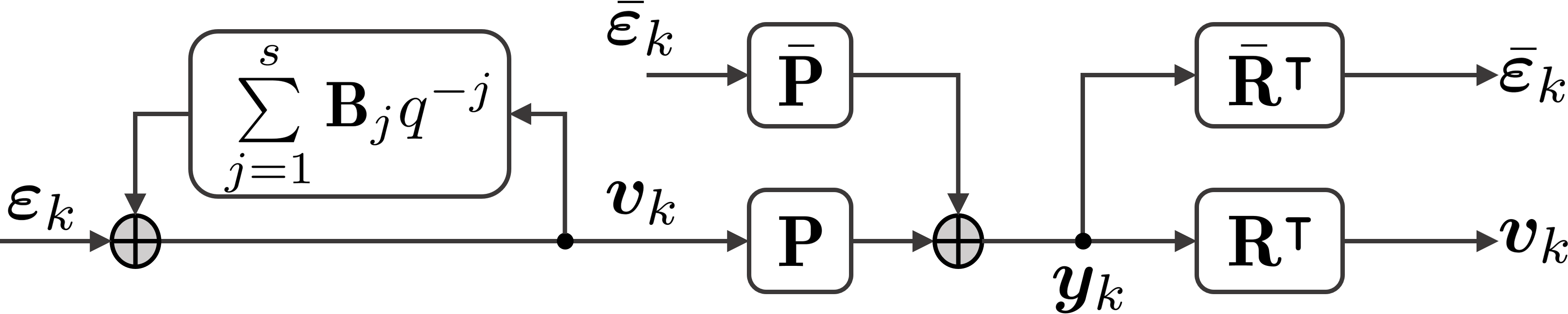}
    \caption{A block diagram of a \modelname~model.}
    \label{fig:block_diag}
\end{figure}

In this paper, we define a {\em probabilistic reduced-dimensional vector autoregressive}  series as
\begin{align}
    &\bm y_k = \mathbf{P} \bm v_k + \mathbf{\bar{P}} \bm{\bar \varepsilon}_k,  &\bm{\bar \varepsilon}_k \sim \mathcal{N}(\bm 0, \mathbf{\Sigma}_{\bm{\bar \varepsilon}}),\label{eq:outer_model}\\
    &\bm v_k = \sum_{j=1}^s\mathbf{B}_j\bm v_{k-j} + \bm{\varepsilon}_k,~~ &\bm{\varepsilon}_k \sim \mathcal{N}(\bm 0, \mathbf{\Sigma}_{\bm{\varepsilon}}),\label{eq:DLV_VAR}
\end{align}
where  $\{\bm{\varepsilon}_k\}$ and~$\{\bm{\bar \varepsilon}_k\}$  are i.i.d. Gaussian noise terms and  $\bm v_k \in \Re^\ell$ denotes the vector of $\ell$ DLVs. Both $\bm{\varepsilon}_j$ and~$\bm{\bar \varepsilon}_j$ form the innovations and therefore are uncorrelated with the past $\bm v_i$ for all~$i<j$. With~$\ell < p$, the loadings matrices~$\mathbf{P} \in \Re^{p \times \ell} $,
$\mathbf{\bar P} \in \Re^{p \times (p-\ell)}$, and $\left[
      \mathbf{P} ~~ \mathbf{\bar{P}}
      \right]\in \Re^{p \times p}$ have full rank. Using the backward-shift operator~$q^{-1}$, the latent dynamic system~\eqref{eq:DLV_VAR} can be converted to
\[
\bm v_k =\left ( \mathbf{I} - \sum_{j=1}^s\mathbf{B}_j q^{-j}
\right )^{-1} \bm{\varepsilon}_k
= \mathbf{G}_v (q^{-1})
 \bm{\varepsilon}_k.
\]
Then, the transfer matrix form of \eqref{eq:outer_model}--\eqref{eq:DLV_VAR} is given by
\begin{align*}
 \bm y_k = \left[
      \mathbf{P}~~ \mathbf{\bar{P}}
      \right]
      \begin{bmatrix}
     \mathbf{G}_v (q^{-1}) & \mathbf{0}  \\
    \mathbf{0} & \mathbf{I}  \end{bmatrix}
     \begin{bmatrix}
     \bm{ \varepsilon}_k   \\ \bm{\bar \varepsilon}_k
      \end{bmatrix}
\end{align*}
which clearly shows the reduced-dimensional dynamics that are commonly observed in routine operational data~\cite{Qin:2022LaVAR_AIChEJ} and economics~\cite{Stock:Watson:2002,gao:tsay:2021high-Dim}. Further, for weakly FDD time series, an RDD approximation is often useful to enable rapid downstream decision-making~\cite {fries2022lasdi}.
It should be noted that the latent VAR model  \eqref{eq:DLV_VAR} can be replaced by a state-space, an ARIMA, or a kernel-based model~\cite{dong:liu:qin:2020DICCA-SVD, yu:qin:2022LaSS,pillonetto2014kernel,khosravi2023existence}.

\subsection{Explicit Expression of DLVs}
The PredVAR model \eqref{eq:outer_model}--\eqref{eq:DLV_VAR} is uniquely characterized by the tuple $\{ \mathbf{P}, \{\mathbf{B}_j \}, \mathbf{\bar P} \}$, but it is implicit in the DLVs~$\bm v_k $. To yield an explicit expression of $\bm v_k $, define $\mathbf{R}\in \Re^{p\times \ell }$ and~$\mathbf{\bar R}\in \Re^{p\times (p-\ell) }$ such that
\begin{equation} \label{eq:RP_relation}
    \left[
      \mathbf{R}~~ \mathbf{\bar{R}}
      \right]^\intercal
    \left[
      \mathbf{P}~~ \mathbf{\bar{P}}
      \right]
    = \mathbf{I},
\end{equation}
which makes~$\left[
      \mathbf{R}~~ \mathbf{\bar{R}}
      \right]$ unique given~$\mathbf{P}$ and~$\mathbf{\bar P}$. The relation \eqref{eq:RP_relation} also implies
    \begin{equation*}
      \mathbf{R}^\intercal \mathbf{ P} = \mathbf{I}, \quad    \mathbf{R}^\intercal  \mathbf{\bar P} = \mathbf{0}, \quad \mathbf{ P}    \mathbf{R}^\intercal +          \mathbf{\bar P}   \mathbf{\bar R}^\intercal= \mathbf{I},
    \end{equation*}
where~$\mathbf{P}\mathbf{R}^\intercal$ and~$\mathbf{\bar P}\mathbf{\bar R}^\intercal$ are two oblique projection matrices because~$(\mathbf{P}\mathbf{R}^\intercal)^2=\mathbf{P}\mathbf{R}^\intercal$ and~$(\mathbf{\bar P}\mathbf{\bar R}^\intercal)^2=\mathbf{\bar P}\mathbf{\bar R}^\intercal$.

The pair $(\mathbf{R}, \mathbf{\bar R}) $
gives explicitly the DLVs~$\bm{v}_k $ and the static noise~$\bm{\bar{\varepsilon}_k}$ from \eqref{eq:outer_model},
\begin{align} \label{eq:vk}
 \bm{v}_k &= \mathbf{R}^\intercal\bm{y}_k,
 \\
    \bm{\bar \varepsilon}_k &= \mathbf{\bar R}^\intercal \bm y_k. \label{eq:bar_epsilonk}
\end{align}
Equations \eqref{eq:vk} and \eqref{eq:bar_epsilonk} jointly explain the dynamic-static decomposition of $\bm y_k$ and reversely
  \eqref{eq:outer_model} gives the composition   of $\bm y_k$. The reversible relations and the \modelname~model are depicted in Fig.~\ref{fig:block_diag} with
 a block diagram. 

\subsection{Identifiability of the \modelname~Model}\label{Sec:identifiability}

 Given any pair of nonsingular matrices~$\mathbf{\bar{M}}\in\Re^{(p-\ell )\times (p-\ell )}$ and~$\mathbf{M}\in\Re^{\ell \times \ell }$, \eqref{eq:outer_model} can be rewritten as
\begin{align}
\bm y_k &= \mathbf{P}  \bm v_k + \mathbf{\bar{P}} \bm{\bar \varepsilon}_k = \mathbf{P} \mathbf{M}^{-1} \mathbf{M} \bm v_k + \mathbf{\bar{P}} \mathbf{\bar M}^{-1} \mathbf{\bar M}  \bm{\bar \varepsilon}_k,    \label{eq:equiv_PredVAR}
\end{align}
which indicates that the tuple~$(\mathbf{P}, \bm v_k, \mathbf{\bar{P}}, \bm{\bar \varepsilon}_k)$  cannot be uniquely identified since it can be exactly represented by~$(\mathbf{P}\mathbf{M}^{-1}, \mathbf{M}\bm v_k, \mathbf{\bar{P}} \mathbf{\bar {M}}^{-1}, \mathbf{\bar {M}}\bm{\bar \varepsilon}_k)$. However, all equivalent realizations differ by similarity transformation only.

 The non-uniqueness of the matrices offers the flexibility to extract desired coordinates or features of the DLVs. For example, we can enforce the following conditions to achieve identifiability.
 \begin{enumerate}
      \item $\mathbf{P}$ orthogonal or the covariance $\mathbf{\Sigma}_{\bm{v}} = \mathbf{I}$ or diagonal, and independently,
     \item $\mathbf{\bar P}$ orthogonal or the covariance $\mathbf{\Sigma}_{\bm{\bar \varepsilon}} = \mathbf{I}$ or diagonal if $\mathbf{\Sigma}_{\bm{\bar \varepsilon}}$ is non-singular.
\item If $\mathbf{\Sigma}_{\bm{\bar \varepsilon}}$ is singular, we can represent
$\bm{\bar \varepsilon}_k = \mathbf{Q} \bm{\bar \varepsilon}_k^* $ with $\mathbf{Q}$ being full-column rank and $\mathbf{\Sigma}_{\bm{\bar \varepsilon}^*} = \mathbf{I}$ having a lower dimension than $\mathbf{\Sigma}_{\bm{\bar \varepsilon}}$.
 \end{enumerate}
In this paper, however, the DLV and static loadings matrices $\mathbf{P}$ and $\mathbf{\bar P}$ need not be mutually orthogonal, requiring an oblique projection for the DLV modeling.

 These options have been used in the literature to enforce diagonal
 $\mathbf{\Sigma}_{\bm{\bar \varepsilon}}$ and~$\mathbf{\Sigma}_{\bm{\varepsilon}}$ as in~\cite{bai2012statistical}, which does not consider rank-deficient innovations, or enforce~$\mathbf{\Sigma}_{\bm{v}}=\mathbf{I}$ as in \cite{Qin:2022LaVAR_AIChEJ}  to achieve the CCA objective. Further, the DLVs in \cite{Qin:2022LaVAR_AIChEJ} are arranged in a non-increasing order of predictability. A rank-deficient
 $\mathbf{\Sigma}_{\bm{\bar \varepsilon}}$ corresponds to a rank-deficient
 $\mathbf{\Sigma}_{\bm{y}}$, which is studied in this paper.
These freedoms allow us to obtain desirable realizations of the \modelname~model.

\subsection{Equivalent Reduced-Rank VAR Models}

A reduced-rank VAR (\rrname)  model is specified as follows: instead of estimating a full-rank coefficient matrix for each lag in a regular VAR model, we impose a factor structure on the coefficient matrices, such that they can be written as a product of lower-rank matrices, namely,
\begin{equation} \label{eq:RRVAR}
    \bm y_k =\sum_{j=1}^s\mathbf{\acute P}\mathbf{\acute B}_j\mathbf{\acute R}^\intercal \bm y_{k-j} +\bm e_k,~~\bm e_k\sim \mathcal{N}(\bm 0, \mathbf{\Sigma}_{\bm{e}}),
\end{equation}
where~$\mathbf{\acute R}^\intercal\mathbf{\acute P}\in\Re^{\ell \times \ell}$ is invertible with $\ell < p$  and~$\bm{e}_k\in \Re^p$ is the full-dimensional innovations vector driving~$\{\bm y_k\}$.

The reduced-rank VAR is analogous to the reduced-rank regression (RRR)  for linear regression \cite{izenman:1975reduced,reinsel:velu:1998}. Just as RRR requires specific solutions that differ from ordinary least squares, \rrname~also requires a specific solution differing from that of regular VAR.  Theorem~\ref{thm:eq_two_models} shows that the \modelname~and \rrname~models are equivalent.

\begin{theorem}\label{thm:eq_two_models}
Equivalent canonical RRVAR of the \modelname~model \eqref{eq:outer_model}.
\begin{enumerate}
    \item The  \modelname~model \eqref{eq:outer_model}--\eqref{eq:DLV_VAR}  is  equivalent  to the following canonical   \rrname~model
\begin{equation} \label{eq:can_RRVAR}
    \bm y_k =\sum_{j=1}^s\mathbf{ P}\mathbf{ B}_j\mathbf{ R}^\intercal \bm y_{k-j} +\bm e_k,~~\bm e_k\sim \mathcal{N}(\bm 0, \mathbf{\Sigma}_{\bm{e}})
\end{equation}
with $\mathbf{ R}^\intercal\mathbf{ P} = \mathbf{ I}$
and the innovation $\bm e_j$  is uncorrelated with $\bm y_i$ for all~$i<j$.
    \item    Each \rrname~model~\eqref{eq:RRVAR}  can be converted to the canonical  \rrname~model~\eqref{eq:can_RRVAR}
   with $\mathbf{ R}^\intercal\mathbf{ P} = \mathbf{ I}$ by setting
   $\mathbf{P} =\mathbf{\acute P} $,
   $ \mathbf{B}_j =\mathbf{\acute B}_j \mathbf{\acute R}^\intercal\mathbf{\acute P}$,
   and
$ \mathbf{R} =\mathbf{\acute R}(\mathbf{\acute P}^\intercal\mathbf{\acute R})^{-1}$.
\end{enumerate}
\end{theorem}

Theorem~\ref{thm:eq_two_models} is proven in Appendix~\ref{apd:eq_two_models}. Since \modelname~and \rrname~represent the same reduced-dimensional dynamics equivalently, one can choose either form for the convenience of the specific analysis and applications.
It is worth noting that $\bm e_k$ is independent of the similarity transforms permitted in \eqref{eq:equiv_PredVAR}.



\section{Optimal Model Estimation Algorithm }\label{sec:alg}

The RRVAR model parameters $(\{\mathbf{B}_j\},\mathbf{P},\mathbf{R}, \mathbf{\Sigma}_{\bm e})$ will be derived based on constrained maximum likelihood estimation. As the parameter matrices appear nonlinearly in the model, the solution algorithm will be iterative, realizing an EM procedure that alternates between the estimations of DLVs and model parameters. To simplify the derivation, we first derive the solution by assuming $\bm{\varepsilon}_k$ and~$\bm{\bar \varepsilon}_k$ are uncorrelated.
The assumption implies that~
\begin{align} \label{eq:Sigma_e}
\left[
      \mathbf{R} ~~ \mathbf{\bar{R}}  \right]^\intercal\mathbf{\Sigma}_{\bm e} \left[
      \mathbf{R} ~~ \mathbf{\bar{R}}  \right] =
\begin{bmatrix}
    \mathbf{\Sigma}_{\bm \varepsilon} & \bm 0\\
    \bm 0 & \mathbf{\Sigma}_{\bm{\bar \varepsilon}}
\end{bmatrix},
 \end{align}
or alternatively,
\begin{align}
\label{eq:Sigmae_inverse}
&\mathbf{\Sigma}_{\bm e}^{-1} \!=\!
\left[
      \mathbf{R} ~~ \mathbf{\bar{R}}  \right]
\!\begin{bmatrix}  \mathbf{\Sigma}_{\bm\varepsilon}^{-1} \\
~ & \mathbf{\Sigma}_{\bm{\bar \varepsilon}}^{-1}
    \end{bmatrix}\!
\left[
      \mathbf{R} ~~ \mathbf{\bar{R}}  \right]^{\intercal}.
\end{align}
In the next section, we show that an optimality condition in the prediction error covariance leads to the uncorrelated noise condition, thus removing the assumption.


\subsection{Necessary Condition for Optimal Solutions}
Given the measured data~$\{\bm y_k \}_{k=1}^{N+s}$, the following likelihood function should be maximized:
\begin{equation*}
     \prod_{k=s+1}^{s+N} p(\bm y_{k}\mid \bm y_{k-1}, \bm y_{k-2}, \ldots,\bm y_{k-s}).
\end{equation*}
Considering the~\rrname~model \eqref{eq:can_RRVAR}, maximizing the likelihood amounts to minimizing the following function constrained by~$\mathbf{R}^\intercal \mathbf{P} = \mathbf{I}$ and~$\mathbf{R}^\intercal \mathbf{\Sigma}_{\bm e}\mathbf{\bar R} = \bm 0$ from \eqref{eq:Sigma_e},
\begin{align}
     &\quad L^{\bm{y}}(\{\mathbf{B}_j\},\mathbf{P},\mathbf{\Sigma}_{\bm e},\mathbf{R}) =N\ln|\mathbf{\Sigma}_{\bm e}| +\sum_{k=s+1}^{s+N}\label{eq:yk_likelihood0} \\
     &\left(\bm y_{k}-\!\!\sum_{j=1}^s\mathbf{P}\mathbf{B}_j\mathbf{R}^\intercal\bm{y}_{k-j}\right)^\intercal \!\!\mathbf{\Sigma}_{\bm e}^{-1}\!\! \left(\bm y_{k}-\!\!\sum_{j=1}^s\mathbf{P}\mathbf{B}_j\mathbf{R}^\intercal\bm{y}_{k-j}\right)\!.\notag
\end{align}
Then, it follows from~\eqref{eq:RP_relation},~\eqref{eq:vk}, and \eqref{eq:Sigmae_inverse} that
\begin{align*}
    & L^{\bm{y}}(\{\mathbf{B}_j\},\mathbf{P},\mathbf{\Sigma}_{\bm{e}},\mathbf{R}) =N\ln|\mathbf{\Sigma}_{\bm{e}}| + \!\!\sum_{k=s+1}^{s+N}\bm{y}_k^\intercal\mathbf{\bar R}\mathbf{\Sigma}_{\bm{\bar \varepsilon}}^{-1}\mathbf{\bar R}^\intercal\bm{y}_k  \nonumber\\
    &+ \sum_{k=s+1}^{s+N}
    \left(\bm{v}_k-\sum_{j=1}^s\mathbf{B}_j\bm{v}_{k-j}\right)^\intercal
    \mathbf{\Sigma}_{\bm{\varepsilon}}^{-1} \left(\bm{v}_k-\sum_{j=1}^s\mathbf{B}_j\bm{v}_{k-j}\right). \label{eq:yk_likelihood2}
\end{align*}
For each~$j\in \enumbracket{s}$, differentiating~$L^{\bm{y}}(\{\mathbf{B}_j\},\mathbf{P},\mathbf{\Sigma}_{\bm e},\mathbf{R})$ with respect to~$\mathbf{B}_j$ and setting it to zero lead to
\begin{multline*}
    \mathbf{B}_j= \sum_{k=s+1}^{s+N}\Bigg(\bm v_{k}\bm v_{k-j}^\intercal-\!\!\sum_{i\in\enumbracket{s},i\neq j}\!\mathbf{B}_i\bm v_{k-i}\bm v_{k-j}^\intercal\Bigg)\times\\
    \hspace{12em}
    \Bigg(\sum_{k=s+1}^{s+N} \bm v_{k-j} \bm v_{k-j}^\intercal \Bigg)^{-1}.
\end{multline*}
Rearranging the above equation gives
\begin{equation} \label{eq:normal_Bj}
 \sum_{i\in\enumbracket{s}} \mathbf{B}_i
\!\sum_{k=s+1}^{s+N}\! \bm v_{k-i}\bm v_{k-j}^\intercal=\!\! \sum_{k=s+1}^{s+N}\! \bm v_{k} \bm v_{k-j}^\intercal, \quad j\in\enumbracket{s}.
\end{equation}
On the other hand, denoting the one-step prediction~
$$\bm{\hat{v}}_k = \sum_{j=1}^s\mathbf{B}_j\bm{v}_{k-j},$$
it follows that \eqref{eq:yk_likelihood0} can be rearranged  as
\begin{multline}\label{eq:yk_likelihood1}
    L^{\bm{y}}(\{\mathbf{B}_j\},\mathbf{P},\mathbf{\Sigma}_{\bm{e}},\mathbf{R})=N\ln|\mathbf{\Sigma}_{\bm{e}}|+\\
    \sum_{k=s+1}^{s+N}(\bm y_{k}-\mathbf{P}\bm{\hat v}_{k})^\intercal \mathbf{\Sigma}_{\bm e}^{-1} (\bm y_{k}-\mathbf{P}\bm{\hat v}_{k}).
\end{multline}
Differentiating \eqref{eq:yk_likelihood1} with respect to $\mathbf{P}$ and~$\mathbf{\Sigma}_{\bm e}^{-1}$  and setting them to zero lead to
\begin{align}
    &\mathbf{P}= \left ( \sum_{k=s+1}^{s+N}\bm y_{k} \bm{\hat v}_{k}^\intercal \right)
    \left(\sum_{k=s+1}^{s+N}\bm{\hat v}_{k}\bm{\hat v}_{k}^\intercal\right)^{-1}, \label{eq:P_condition} \\
    &\mathbf{\Sigma}_{\bm e} = \frac{1}{N}\sum_{k=s+1}^{s+N}(\bm y_{k}-\mathbf{P}\bm{\hat v}_{k})(\bm y_{k}-\mathbf{P}\bm{\hat v}_{k})^\intercal.
    \label{eq:Sigma_e_condition}
\end{align}
Substituting~\eqref{eq:P_condition} into~\eqref{eq:Sigma_e_condition} leads to
\begin{multline}\label{update_e_condition}
    \mathbf{\Sigma}_{\bm e}= \frac{1}{N}\Bigg(\sum_{k=s+1}^{s+N}\bm y_{k}\bm y_{k}^\intercal-\\
    \!\!\left ( \sum_{k=s+1}^{s+N}\bm y_{k} \bm{\hat v}_{k}^\intercal \right)
    \left(\sum_{k=s+1}^{s+N}\bm{\hat v}_{k}\bm{\hat v}_{k}^\intercal\right)^{-1}\!\!\!\left ( \sum_{k=s+1}^{s+N} \bm{\hat v}_{k}\bm y_{k}^\intercal \right)\Bigg).
\end{multline}

\subsection{The EM Solution }

It is clear that the necessary condition of optimality involves nonlinear equations of the parameters $ (\{\mathbf{B}_j\},\mathbf{P},\mathbf{\Sigma}_{\bm e},\mathbf{R})$. To solve them iteratively, we separate the solution into two EM steps. First,
given $\mathbf{\hat R}$ and $\{\mathbf{\hat  B}_j\}$ respectively, the E-step of the EM procedure gives
\begin{align}
   & \bm{ v}_{k|\mathbf{\hat R}}  = \mathbf{\hat R}^\intercal \bm{y}_k \text{, and } \label{update_vk}\\
&\bm{\hat{{v}}}_{k|\mathbf{\hat R}}  = \sum_{j=1}^s\mathbf{\hat B}_j\bm{ v}_{k-j|\mathbf{\hat R}}, \label{update_vk_prediction}
\end{align}
 for $k=1, 2, \cdots, N+s$.

Second, \eqref{update_vk} and \eqref{update_vk_prediction} can be used to estimate the model parameters in the M-step. Formulate matrices
\begin{align*}
    \mathbf{Y}_i & \triangleq~ [\bm y_{i+1}~\bm y_{i+2}~\cdots~\bm y_{i+N}]^{\intercal}, i\in \{0\}\cup\enumbracket{s};\\
     \mathbf{V}_i &\triangleq~ \left[{\bm v}_{i+1|\mathbf{\hat R}}~{\bm v}_{i+2|\mathbf{\hat R}}~\cdots~{\bm v}_{i+N|\mathbf{\hat R}}\right]^{\intercal}, i\in \{0\}\cup\enumbracket{s};\\
     & =~ \mathbf{Y}_i \mathbf{\hat R};\\
     {\mathbb{V}}~&\triangleq~ [{\mathbf{V}}_{s-1}~{\mathbf{V}}_{s-2}~\cdots~{\mathbf{V}}_0]; \\
     {\mathbb{\hat B}}~&\triangleq~  [{\mathbf{\hat B}}_1~{\mathbf{\hat B}}_2~\cdots~{\mathbf{\hat B}}_s]^{\intercal},
\end{align*}
which  include the formation of $\mathbf{ V}_s$ and $\mathbf{Y}_s$. Substituting the estimated DLV vectors \eqref{update_vk} into
\eqref{eq:normal_Bj} leads to
\begin{equation}
   {\mathbb{V}}^{\intercal}{\mathbb{V}} \hat{\mathbb{B}} = {\mathbb{V}}^{\intercal}\mathbf{ V}_{s} ~~\Longrightarrow~~ \hat{\mathbb{B}} = ({\mathbb{V}}^{\intercal}{\mathbb{V}})^{-1}{\mathbb{V}}^{\intercal}\mathbf{ V}_{s}.
    \label{update_B}
\end{equation}
Further,  denoting
\begin{align*}
         \mathbf{\hat V}_s&\triangleq~ \left[\hat {\bm v}_{s+1|\mathbf{\hat R}}~\hat {\bm v}_{s+2|\mathbf{\hat R}}~\cdots~\hat{\bm v}_{s+N|\mathbf{\hat R}}\right]^{\intercal},
\end{align*}
 \eqref{update_vk_prediction} and~\eqref{update_B} lead to~
\begin{equation}\label{update_Vs_hat}
\mathbf{\hat V}_s= {\mathbb{V}}\hat{\mathbb{B}} ={\mathbb{V}} ({\mathbb{V}}^{\intercal}{\mathbb{V}})^{-1}{\mathbb{V}}^{\intercal} \mathbf{ V}_{s} = \mathbf{\Pi}_{\mathbb{ V}}  \mathbf{ V}_{s},
\end{equation}
where $\mathbf{\Pi}_{\mathbb{ V}} \triangleq {\mathbb{V}} ({\mathbb{V}}^{\intercal}{\mathbb{V}})^{-1}{\mathbb{V}}^{\intercal}$ is a projection matrix,
which shows that $\mathbf{\hat V}_s$ is  the orthogonal projection of~$\mathbf{V}_s$ onto the range of $\mathbb{V}$. It follows from  \eqref{eq:P_condition} and \eqref{update_e_condition} that
\begin{align}
  {\mathbf{\hat P}} &=\mathbf{Y}_s^{\intercal} \mathbf{\hat V}_s (\mathbf{\hat V}_s^{\intercal} \mathbf{\hat V}_s)^{-1}, \label{update_P}\\
  \mathbf{\hat \Sigma}_{\bm e} &=  \mathbf{Y}_s^{\intercal}\mathbf{Y}_s/N- \mathbf{Y}_s^\intercal \mathbf{\hat V}_s (\mathbf{\hat V}_s^{\intercal} \mathbf{\hat V}_s)^{-1}\mathbf{\hat V}_s^\intercal \mathbf{Y}_s/N \nonumber \\
 &= \mathbf{Y}_s^\intercal (\mathbf{I} -\mathbf{\Pi}_{\mathbf{\hat V}_s}) \mathbf{Y}_s/N, \label{update_Sigma_e}
\end{align}
where $\mathbf{\Pi}_{\mathbf{\hat V}_s} \triangleq  \mathbf{\hat V}_s (\mathbf{\hat V}_s^{\intercal} \mathbf{\hat V}_s)^{-1}\mathbf{\hat V}_s^\intercal  $ is a projection matrix.

Finally, we need to update $\mathbf{\hat R}$ to complete the iteration loop. Based on \eqref{eq:Sigma_e}, pre-multiplying $ \mathbf{\hat R}^\intercal $ and post-multiplying $\mathbf{\hat R}$ to \eqref{update_Sigma_e} lead to
\begin{align}
 \mathbf{\hat \Sigma}_{\bm \varepsilon} &=
 \mathbf{\hat R}^\intercal \mathbf{\hat \Sigma}_{\bm e}  \mathbf{\hat R} =
\mathbf{\hat R}^\intercal \mathbf{Y}_s^\intercal (\mathbf{I} -\mathbf{\Pi}_{\mathbf{\hat V}_s}) \mathbf{Y}_s \mathbf{\hat R}/N \nonumber  \\
&= \mathbf{V}_s^\intercal \mathbf{V}_s/N -
\mathbf{V}_s^\intercal \mathbf{\Pi}_{\mathbf{\hat V}_s} \mathbf{V}_s/N \label{eq:Sigma_eps}
\end{align}
From the orthogonal projection in \eqref{update_Vs_hat}, it is straightforward to show that
$\mathbf{\hat V}_s^\intercal (\mathbf{ V}_s - \mathbf{\hat V}_s) = \mathbf{0}$, which gives
$\mathbf{\hat V}_s^\intercal \mathbf{ V}_s = \mathbf{\hat V}_s^\intercal \mathbf{\hat V}_s$. The last term in \eqref{eq:Sigma_eps} becomes
\begin{align*}
    \mathbf{V}_s^\intercal \mathbf{\Pi}_{\mathbf{\hat V}_s} \mathbf{V}_s/N
    &= \mathbf{V}_s^\intercal \mathbf{\hat V}_s (\mathbf{\hat V}_s^{\intercal} \mathbf{\hat V}_s)^{-1}\mathbf{\hat V}_s^\intercal \mathbf{V}_s/N \\
   & = \mathbf{\hat V}_s^{\intercal} \mathbf{\hat V}_s/N
    \triangleq~ \mathbf{\hat \Sigma}_{\hat{\bm v}}
\end{align*}
Further, \eqref{eq:Sigma_eps} becomes
\begin{align}
  \mathbf{\hat \Sigma}_{\bm \varepsilon} =  \mathbf{\hat \Sigma}_{\bm{v}} - \mathbf{\hat \Sigma}_{\hat{\bm v}} \label{eq:Sigma_relation}
\end{align}
The best predictive DLV model is one that has the smallest prediction error covariance $\mathbf{\hat \Sigma}_{\bm \varepsilon}$ given  $\mathbf{\hat R}$
in the sense that $\mathbf{\hat \Sigma}'_{\bm \varepsilon} -
\mathbf{\hat \Sigma}_{\bm \varepsilon}$ is positive semi-definite for any other $\mathbf{\hat \Sigma}'_{\bm \varepsilon}$.
However, since
$\mathbf{\hat \Sigma}_{\bm \varepsilon}$
 can be made arbitrarily small by scaling $\mathbf{\hat R}$, we need to fix the scale of one of the terms in \eqref{eq:Sigma_relation}. In \cite{Qin:2022LaVAR_AIChEJ}, the scale is fixed by enforcing $\mathbf{V}_s^\intercal \mathbf{V}_s     = \mathbf{I}$. In this paper, an equivalent scaling is implemented in the next subsection with
 \begin{align}
   \mathbf{\hat \Sigma}_{\bm v} =  \mathbf{V}_s^\intercal \mathbf{V}_s /N=\mathbf{\hat R}^\intercal \mathbf{\hat \Sigma}_{\bm y}\mathbf{\hat R}     = \mathbf{I},  \label{eq:Sigma_v=I}
 \end{align}
 where~$\mathbf{\hat \Sigma}_{\bm y}= \mathbf{Y}_s^\intercal  \mathbf{Y}_s /N$. Then \eqref{eq:Sigma_relation} becomes
\begin{align*}
 \mathbf{\hat \Sigma}_{\bm \varepsilon} =  \mathbf{I} - \mathbf{\hat \Sigma}_{\hat{\bm v}}
\end{align*}
Therefore, maximizing
$\mathbf{\hat \Sigma}_{\hat{\bm v}}  = \mathbf{\hat R}^\intercal   \mathbf{Y}_s^\intercal \mathbf{\Pi}_{\mathbf{\hat V}_s} \mathbf{Y}_s \mathbf{\hat R}/N$
subject to \eqref{eq:Sigma_v=I} leads to the smallest $\mathbf{\hat \Sigma}_{\bm \varepsilon}$.

\subsection{An Optimal \modelname~Algorithm with Normalization}
The constraint \eqref{eq:Sigma_v=I} can be implemented by normalizing the data based on $\mathbf{\hat \Sigma}_{\bm y}$, a technique used in \cite{Qin:2022LaVAR_AIChEJ,gao:tsay:2021high-Dim}. The constraint \eqref{eq:Sigma_v=I} indicates that $\mathbf{\hat R}$ can only be found in the eigenspace of $\mathbf{\hat \Sigma}_{\bm y}$ corresponding to the non-zero eigenvalues. Thus, assuming $\textrm{rank}(\mathbf{\hat \Sigma}_{\bm y}) = r \leq p$, we must have the number of DLVs $\ell \leq r$.

By performing eigenvalue decomposition (EVD) on~
\begin{equation*}\label{eq:EVD_Sigma_y}
   \mathbf{\hat \Sigma}_{\bm y}
    = \frac{\mathbf{Y}_s^\intercal  \mathbf{Y}_s}{N}
    =
[\mathbf{U}~~ \mathbf{\tilde U}]
\begin{bmatrix}
  \mathbf{D} & \mathbf{0} \\
 \mathbf{0}  & \mathbf{0}
\end{bmatrix}
[\mathbf{U}~~ \mathbf{\tilde U}]^\intercal
=\mathbf{U}\mathbf{D}\mathbf{U}^\intercal,
\end{equation*}
where~$\mathbf{U} \in \Re^{p \times r}$,
$\mathbf{\tilde U} \in \Re^{p \times (p-r)}$, and $\mathbf{D} \in \Re^{r \times r}$ contains the non-zero eigenvalues in non-increasing order, we define a normalized data vector $\bm y_k^* \in \Re^{r}$ with
\begin{align} \label{eq:yk=UD yk*}
 \bm{ y}_k = \mathbf{U} \mathbf{D}^{\frac{1}{2}} \bm y^*_k,
\end{align}
 which represents~$\bm y_k^*$ with the following normalized PredVAR model
\begin{align}
    \bm{ y}^*_k = \mathbf{D}^{-\frac{1}{2}} \mathbf{U}^\intercal \bm y_k
= \mathbf{P}^* \bm{ v}_k + \mathbf{\bar P}^* \bm{\bar \varepsilon}_k^*, \label{eq:PredVAR_yk*}
\end{align}
where $\mathbf{P}^*  \in \Re^{r \times \ell}$,
$\mathbf{\bar P}^*  \in \Re^{r \times (r - \ell)}$, and we
make $ \mathbf{\Sigma}_{\bm{\bar \varepsilon}^*} = \mathbf{I}$ without any loss of generality. Then, it follows from \eqref{eq:yk=UD yk*} and \eqref{eq:PredVAR_yk*} that
\begin{align} \label{eq:P=UD P*}
\mathbf{P}
   = \mathbf{U} \mathbf{D}^{\frac{1}{2}}  \mathbf{P}^*
\end{align}
Forming $ \mathbf{Y}_i^*$ the same way as $ \mathbf{Y}_i$ for $i= 0, 1, \cdots, s$, it is straightforward to show that
$ \mathbf{\hat \Sigma}_{\bm y^*} = \mathbf{Y}_s^{*\intercal}  \mathbf{Y}_s^*/N = \mathbf{I}$.
 By using the uncorrelated condition of the DLVs and static noise, we have
\begin{align}
  \mathbf{I} &=
 \mathbf{\hat \Sigma}_{\bm y^*} \!=\!
  \mathbf{P}^*  \mathbf{\hat \Sigma}_{\bm v} \mathbf{P}^{* \intercal} \!+\!
  \mathbf{\bar P}^*
    \mathbf{\hat \Sigma}_{\bm{\bar \varepsilon}^*}
    \mathbf{\bar P}^{* \intercal} =
  \mathbf{P}^*   \mathbf{P}^{* \intercal} +
  \mathbf{\bar P}^*
    \mathbf{\bar P}^{* \intercal} \nonumber \\
   & =
      \left[
        \mathbf{P}^*~~ \mathbf{\bar P}^*
    \right]
        \left[
        \mathbf{P}^* ~~ \mathbf{\bar P}^*
    \right]^\intercal  =
      \left[
        \mathbf{P}^* ~~ \mathbf{\bar P}^*
    \right] ^\intercal
        \left[
        \mathbf{P}^* ~~ \mathbf{\bar P}^*
    \right],
    \label{eq:P*_Pbar*}
\end{align}
i.e.,
$\mathbf{P}^{*\intercal} \mathbf{P}^* = \mathbf{I} $,
$\mathbf{\bar P}^{*\intercal} \mathbf{\bar P}^* = \mathbf{I} $, and
$\mathbf{P}^{*\intercal} \mathbf{\bar P}^* = \mathbf{0} $. Comparing the above relation to \eqref{eq:RP_relation}, it is clear that $ \mathbf{P}^*$ and $ \mathbf{R}^*$ coincide with each other, i.e., $ \mathbf{R}^* = \mathbf{P}^*$. Therefore, we generate the DLVs with
\begin{equation*} \label{eq_vk_star}
 \bm{ v}_k  ={\mathbf{R}^*}^\intercal \bm{ y}^*_k
={\mathbf{R}^*}^\intercal\mathbf{D}^{-\frac{1}{2}} \mathbf{U}^\intercal \bm y_k
= {\mathbf{R}}^\intercal \bm{ y}_k,
\end{equation*}
giving $\mathbf{R} = \mathbf{U}\mathbf{D}^{-\frac{1}{2}} \mathbf{R}^* $. Therefore, the optimal PredVAR solution is converted to finding
 $\mathbf{R}^* $, and thus  $\mathbf{R} = \mathbf{U}\mathbf{D}^{-\frac{1}{2}} \mathbf{R}^* $, to make $\bm{ v}_k$ most predictable.

To find the estimate $\mathbf{\hat R}^*$, the constraint \eqref{eq:Sigma_v=I} becomes
 \begin{align}
   \mathbf{\hat \Sigma}_{\bm v} =
\mathbf{\hat R}^{*\intercal} \left(\mathbf{Y}_s^{*\intercal}  \mathbf{Y}_s^*/N\right) \mathbf{\hat R}^*     =
\mathbf{\hat R}^{*\intercal}  \mathbf{\hat R}^*     = \mathbf{I},  \label{eq:Sigma_v=I*}
 \end{align}
which again makes $\mathbf{\hat R}^*$ an orthogonal matrix.
 Equation
\eqref{eq:Sigma_eps} becomes
\begin{equation}
  \mathbf{\hat \Sigma}_{\bm \varepsilon} =  \mathbf{I} - \mathbf{\hat \Sigma}_{\hat{\bm v}} = \mathbf{I}
  - \mathbf{\hat R}^{* \intercal} \left( \mathbf{Y}_s^{* \intercal} \mathbf{\Pi}_{\mathbf{\hat V}_s} \mathbf{Y}_s^*/N\right) \mathbf{\hat R}^*.
  \label{eq:Sigma_v hat}
\end{equation}
Therefore, to minimize $ \mathbf{\hat \Sigma}_{\bm \varepsilon}$  subject to \eqref{eq:Sigma_v=I*} with $\ell$ DLVs, $\mathbf{\hat R}^*$ must contain the eigenvectors of the $\ell$ largest eigenvalues of
$ \mathbf{Y}_s^{* \intercal} \mathbf{\Pi}_{\mathbf{\hat V}_s} \mathbf{Y}_s^*/N$.  Performing EVD on~
\begin{align} \label{eq:EVD_PredVAR}
\mathbf{ J} =
\mathbf{Y}_s^{* \intercal} \mathbf{\Pi}_{\mathbf{\hat V}_s} \mathbf{Y}_s^*/N =\mathbf{W}\mathbf{\Lambda} \mathbf{W}^\intercal,
\end{align}
where~$\mathbf{\Lambda} $  contains the eigenvalues in non-increasing order, it is convenient to choose~
\begin{align} \label{eq:R*=P*solution}
 \mathbf{\hat R}^*=\mathbf{\hat P}^*=\mathbf{W}(:,1:\ell)
\end{align}
as the optimal solution,  making
\begin{equation*} \label{eq:Sigma_hat_v}
    \mathbf{\hat \Sigma}_{\hat{\bm v}}  = \mathbf{\Lambda}(1:\ell,1:\ell)
    =
  \textrm{diag} (\lambda_1,\lambda_2, \cdots, \lambda_\ell  )
\end{equation*}
contain the leading canonical correlation coefficients (i.e., $R^2$ values) of the DLVs
and
\begin{equation} \label{eq:Sigma_epsilon}
  \mathbf{\hat \Sigma}_{\bm \varepsilon} =  \mathbf{I} -  \mathbf{\Lambda}(1:\ell,1:\ell)
\end{equation}
the residual covariance. It is noted that if some diagonal elements of $\mathbf{\Lambda}(1:\ell,1:\ell)$ are zero, the noise covariance $\mathbf{\hat \Sigma}_{\bm \varepsilon}$ is rank-deficient, and the corresponding DLVs have perfect predictions.

 To summarize, the whole EM iteration solution boils down to updating $\mathbf{\hat R}^*$ and $\mathbb{\hat B}$ until convergence. With the converged  $\mathbf{\hat R}^*$  and  $\mathbb{\hat B}$, we can calculate $\mathbf{\hat V}_s$, $\mathbf{\hat R}$,
 $\mathbf{\hat P}$ from \eqref{eq:P=UD P*}, and $\mathbf{\hat \Sigma}_{\bm e}$ from \eqref{update_Sigma_e}.

The static part of the model can be calculated from the oblique complement of the dynamic counterpart. It is straightforward to calculate
$\mathbf{\hat {\bar P}^*}$ from \eqref{eq:P*_Pbar*}. From \eqref{eq:yk=UD yk*} and \eqref{eq:PredVAR_yk*}, we have
\begin{align*}
\bm{ y}_k &= \mathbf{U} \mathbf{D}^{\frac{1}{2}} \bm y^*_k  +
\mathbf{\tilde U} \cdot \mathbf{0} \\
&=
\mathbf{U} \mathbf{D}^{\frac{1}{2}} \mathbf{P}^* \bm{ v}_k +
\mathbf{U} \mathbf{D}^{\frac{1}{2}} \mathbf{\bar P}^* \bm{\bar \varepsilon}_k^*
 +
\mathbf{\tilde U} \cdot \mathbf{0} \\
&=
\mathbf{U} \mathbf{D}^{\frac{1}{2}} \mathbf{P}^* \bm{ v}_k +
[\mathbf{U} \mathbf{D}^{\frac{1}{2}} \mathbf{\bar P}^*  ~~\mathbf{\tilde U} ]
\begin{bmatrix}
   \bm{\bar \varepsilon}_k^* \\
   \mathbf{0}
\end{bmatrix}
\end{align*}
Therefore, to match \eqref{eq:outer_model} and  satisfy \eqref{eq:RP_relation}, we can choose
\begin{align}
    \mathbf{\hat {\bar P}} &=[\mathbf{U} \mathbf{D}^{\frac{1}{2} }\mathbf{\hat {\bar P}^*} ~~
    \mathbf{\tilde U}] \text{ and }\label{eq:Pbar_solution} \\
    \mathbf{\hat {\bar R}} &=[\mathbf{U} \mathbf{D}^{-\frac{1}{2}}\mathbf{\hat {\bar R}^*}~~ \mathbf{\tilde U}]
    \label{eq:Rbar_solution}
\end{align}
 which  give
the static noise
\begin{align} \label{eq:epsilon_bar}
    {\bm {\bar \varepsilon}}_k =
    \begin{bmatrix}
   \bm{\bar \varepsilon}_k^* \\
   \mathbf{0}
\end{bmatrix}
= \mathbf{\hat {\bar R}}^\intercal
\bm {y}_k \quad
\text{with} \quad
\mathbf{\hat \Sigma}_{\bm {\bar \varepsilon}} =
\begin{bmatrix}
  \mathbf{\bar D}^2 & \\
  & \mathbf{0}
\end{bmatrix}.
\end{align}

The pseudocode of the \modelname~algorithm is shown in Algorithm~\ref{alg:PredVAR}. The data are first normalized based on~$\mathbf{\hat \Sigma}_{\bm{y}}$ before the EM iteration. Then, the estimates $(\mathbf{\hat R^*}, \mathbb{\hat B})$ are iterated with the EM-based \modelname~solution. After convergence, we
obtain $(\{\mathbf{\hat  B}_j\}, \mathbf{\hat P}, \mathbf{\hat R}, \mathbf{\hat \Sigma}_{\bm{e}})$ for the RRVAR model and $(\mathbf{\hat {\bar P} },\mathbf{\hat  {\bar R}}, \mathbf{\hat \Sigma}_{\bm{\bar \varepsilon}})$ for the static noise model.

\subsection{Rank-deficient Covariances of Innovations }

In real applications, it is possible that the covariances $\mathbf{\hat \Sigma}_{\bm y}$, $\mathbf{\hat \Sigma}_{\bm \varepsilon}$, and $\mathbf{\hat \Sigma}_{\bm{\bar \varepsilon}} $ can be rank-deficient. Algorithm \ref{alg:PredVAR} handles the rank deficiency of $\mathbf{\hat \Sigma}_{\bm y}$ by the initial EVD step. Since $\mathbf{\hat \Sigma}_{\bm v} = \mathbf{I}$ is enforced in this algorithm, the null space of $\mathbf{\hat \Sigma}_{\bm y}$ cannot be part of $\bm v_k$, and therefore, it must be part of $\mathbf{\hat \Sigma}_{\bm{\bar \varepsilon}}$. On the other hand, with the enforced $\mathbf{\hat \Sigma}_{\bm v} = \mathbf{I}$, it is possible to have rank-deficient $\mathbf{\hat \Sigma}_{\bm \varepsilon}$, which corresponds to DLVs perfectly predictable. Therefore, we give the following remarks.

\begin{remark}
Singularities in  $\mathbf{\hat \Sigma}_{\bm \varepsilon}$ are considered in the PredVAR model as the leading DLVs that are perfectly predictable. Let
\begin{align*}
    {\bm \varepsilon}_k = \begin{bmatrix}
 {\bm \varepsilon}_k^{ \circ } \\ {\bm \varepsilon}_k^{1}
    \end{bmatrix}
\end{align*}
where ${\bm \varepsilon}_k^{ \circ } = \mathbf{0}$ corresponds to all perfectly predictable DLVs. Similarly partitioning
$\mathbf{P} = [\mathbf{P}^\circ ~~ \mathbf{P}^1 ] $, we have the following relation from \eqref{eq:can_RRVAR}
\begin{align} \label{eq:RR_innovation_RRVAR}
 \bm y_k =\sum_{j=1}^s\mathbf{ P}\mathbf{ B}_j\mathbf{ R}^\intercal \bm y_{k-j} +
 [\mathbf{P}^1 ~~ \mathbf{\bar P} ]
  \begin{bmatrix}
 {\bm \varepsilon}_k^{1 } \\ \bar{\bm  \varepsilon}_k
    \end{bmatrix}.
\end{align}
For the case of full rank $\mathbf{\hat \Sigma}_{\bm y}$, \eqref{eq:RR_innovation_RRVAR} gives a VAR model with the rank-reduced innovation $  \begin{bmatrix}
 {\bm \varepsilon}_k^{1 } \\ \bar{\bm  \varepsilon}_k
    \end{bmatrix}$ with a dimension less than that of $\bm y_k$. The situation of rank-reduced innovations is a topic of active research, e.g., \cite{WEERTS:VandenHof:2018RR_noise} and~\cite{cao2023identification}.
\end{remark}

\begin{remark}
For the case of  rank-deficient $\mathbf{\hat \Sigma}_{\bm y}$, its null space is contained in $\mathbf{\hat \Sigma}_{\bm{\bar \varepsilon}}$,  making $\mathbf{\hat \Sigma}_{\bm{\bar \varepsilon}}$ rank-deficient.
Therefore, rank-deficient  $\mathbf{\hat \Sigma}_{\bm{ \varepsilon}}$ and  $\mathbf{\hat \Sigma}_{\bm{\bar \varepsilon}}$
are naturally taken care of in Algorithm \ref{alg:PredVAR}.
\end{remark}

\begin{algorithm}[t]
\caption{An Optimal \modelname~Algorithm.}\label{alg:PredVAR}
\LinesNumbered
\SetNoFillComment
\KwIn{$\ell$; $s$; zero-centering measurements $\{\bm y_k\}_{k=1}^{N+s}$; }
  \KwOut{$\mathbf{\hat B}_j,j\in \enumbracket{s}$;~$\mathbf{\hat P}$, $\mathbf{\hat \Sigma}_{\bm e}$, $\mathbf{\hat R}$;  $\mathbf{\hat {\bar P} },\mathbf{\hat \Sigma}_{\bm{\bar \varepsilon}},\mathbf{\hat  {\bar R}}$;}

  {Normalization: perform EVD on~$\mathbf{Y}_s^{\intercal}\mathbf{Y}_s/{N}\!=\!\mathbf{U}\mathbf{D}\mathbf{U}^\intercal$\!,} {where $\mathbf{D}$ contains non-zero eigenvalues in non-increasing order; $\mathbf{Y}^*=\mathbf{Y}\mathbf{U}\mathbf{D}^{-\frac{1}{2}}$\;}
  {Initialize~$\hat{\mathbf{{R}}}^*$\;}
  \While {the convergence condition is unsatisfied}{
  {Update $\mathbf{V}_i = \mathbf{Y}_i^*\mathbf{\hat R}^{*}, i\in\{0\}\cup \enumbracket{s}$ and $\mathbb{V}$\; \label{alg:updatevk}}
  {Update~$\hat{\mathbb{B}} = ({\mathbb{V}}^{\intercal}{\mathbb{V}})^{-1}{\mathbb{V}}^{\intercal}\mathbf{ V}_{s}$ and $\mathbf{\hat V}_s =\mathbb{V} \hat{\mathbb{B}}$\;}
  {Update $ \mathbf{ J} = \mathbf{Y}_s^{*\intercal} \mathbf{\hat V}_s (\mathbf{\hat V}_s^{\intercal} \mathbf{\hat V}_s)^{-1}\mathbf{\hat V}_s ^\intercal \mathbf{Y}^*_s/N$\;\label{alg:J}}
  {Perform EVD on $ \mathbf{ J} =\mathbf{W}\mathbf{\Lambda}\mathbf{W}^\intercal$ where~$\mathbf{\Lambda}$ is} {diagonal containing the eigenvalues in non-increasing order; $\mathbf{\hat R}^*=\mathbf{W}(:,1:\ell)$\; \label{update_R}}
  }
{$\mathbf{\hat \Sigma}_{\bm{\varepsilon}} =\mathbf{I}- \mathbf{\Lambda}(1:\ell,1:\ell)$\;}
{De-normalization: $\mathbf{\hat R}=\mathbf{U}\mathbf{D}^{-\frac{1}{2}} \mathbf{\hat R}^*$ and  calculate~$\mathbf{\hat P}$  from \eqref{eq:P=UD P*}, $\mathbf{\hat \Sigma}_{\bm{ \varepsilon}}$ from \eqref{eq:Sigma_epsilon},} and $\mathbf{\hat \Sigma}_{\bm e}$ from \eqref{update_Sigma_e}\;
{Calculate the static noise model $(\mathbf{\hat {\bar P} },\mathbf{\hat  {\bar R}}, \mathbf{\hat \Sigma}_{\bm{\bar \varepsilon}})$ base on \eqref{eq:Pbar_solution}, \eqref{eq:Rbar_solution},  and \eqref{eq:epsilon_bar}.
 }
\end{algorithm}

\subsection{Initialization}

We give the following effective strategy for initializing~$\hat{\mathbf{{R}}}^*$, which is consistent with the updating of~$\hat{\mathbf{{R}}}^*$ in Algorithm~\ref{alg:PredVAR}. Specifically, similar to~$\mathbb{V}$, form the augmented  matrix
$$
{\mathbb{Y}^*} ~\triangleq~ [\mathbf{Y}^*_{s-1}~\mathbf{Y}^*_{s-2}~\cdots~\mathbf{Y}^*_0].
$$
Replacing~$\mathbf{\hat V}_s$ with~$\mathbb{Y}^*$ in \eqref{eq:EVD_PredVAR} and performing EVD give rise to
\begin{align} \label{eq:EVD_initialize}
    \mathbf{ J}_0=\mathbf{Y}_s^{*\intercal} \mathbf{\Pi}_{\mathbb{Y}^*}  \mathbf{Y}^*_s/N  =\mathbf{W}_0\mathbf{\Lambda}_0\mathbf{W}_0^\intercal,
\end{align}
 where
 $\mathbf{\Pi}_{\mathbb{Y}^*}  =  \mathbb{Y}^* ( \mathbb{Y}^{* \intercal} \mathbb{Y}^*)^{-1}  \mathbb{Y}^{* \intercal}$  and
 $\mathbf{\Lambda}_0$ contains the eigenvalues in a non-increasing order, we can initialize~$\mathbf{\hat R}^*$ as~$\mathbf{W}_0(:,1:\ell)$.

The underlying idea is to estimate the error covariance matrix~$\mathbf{\hat \Sigma}_{\bm{e}^f}$ from the following full-rank VAR model
\begin{equation*} 
    \bm{y}_k^* =\sum_{j=1}^s \mathbf{A}_j\bm y^*_{k-j} +\bm{e}_k^f,~~\bm e_k^f\sim \mathcal{N}(\bm 0, \mathbf{\Sigma}_{\bm{e}^f}),
\end{equation*}
which makes
\[
\mathbf{\hat \Sigma}_{\bm{e}^f} =
\mathbf{Y}_s^{*\intercal} (\mathbf{I}-
\mathbf{\Pi}_{\mathbb{Y}^*} ) \mathbf{Y}^*_s/N =
\mathbf{I}-\mathbf{ J}_0.
\]
Interestingly, the canonical analysis of time series in \cite{box:tiao:1977} estimates the VAR error covariance matrix first, and then performs EVD to select $\ell$ canonical components for the smallest eigenvalues of the error covariance matrix, which is equivalent to the selection of $\mathbf{\hat R}^*$ in this initialization step. Therefore, the work\cite{box:tiao:1977} serves as an initial step of the PredVAR iteration only.

\section{Optimal RDD Decomposition} \label{sec:opt_dyn_static_decomp}

While there exists a one-to-one correspondence between the dynamic-static decomposition of $\bm{y}_k$ and the ranges of~$\mathbf{ P}$ and~$\mathbf{\bar P}$, the decomposition is not unique. This fact offers the flexibility to extract DLVs and static noise with desired properties. Particularly, the data series can be decomposed into a dynamic series~$\{\mathbf{P}\bm{v}_k\}$ and a static noise series~$\{\mathbf{\bar P}\bm{\bar \varepsilon}_k\}$ to make
\begin{enumerate}
    \item the DLV series~$\{\bm{v}_k\}$  as predictable or the covariance of the error $\{\bm{ \varepsilon}_k\}$ as small as possible, and
    \item a realization of the static noise series~$\{\bm{\bar \varepsilon}_k\}$ the least correlated with the DLV innovations series $\{\bm{ \varepsilon}_k\}$.
\end{enumerate}
The latter is feasible since the realization of  $\mathbf{\bar P}$ needs not to be orthogonal to $\mathbf{ P}$ in PredVAR, which is the benefit of using oblique projections. Interestingly, our analysis also reveals the consistency between the two goals.

\subsection{Uncorrelated Noise Realization}
\begin{figure*}[t!]
\centering
\begin{subfigure}{.49\textwidth}
  \centering
  \includegraphics[height=0.6\textwidth]{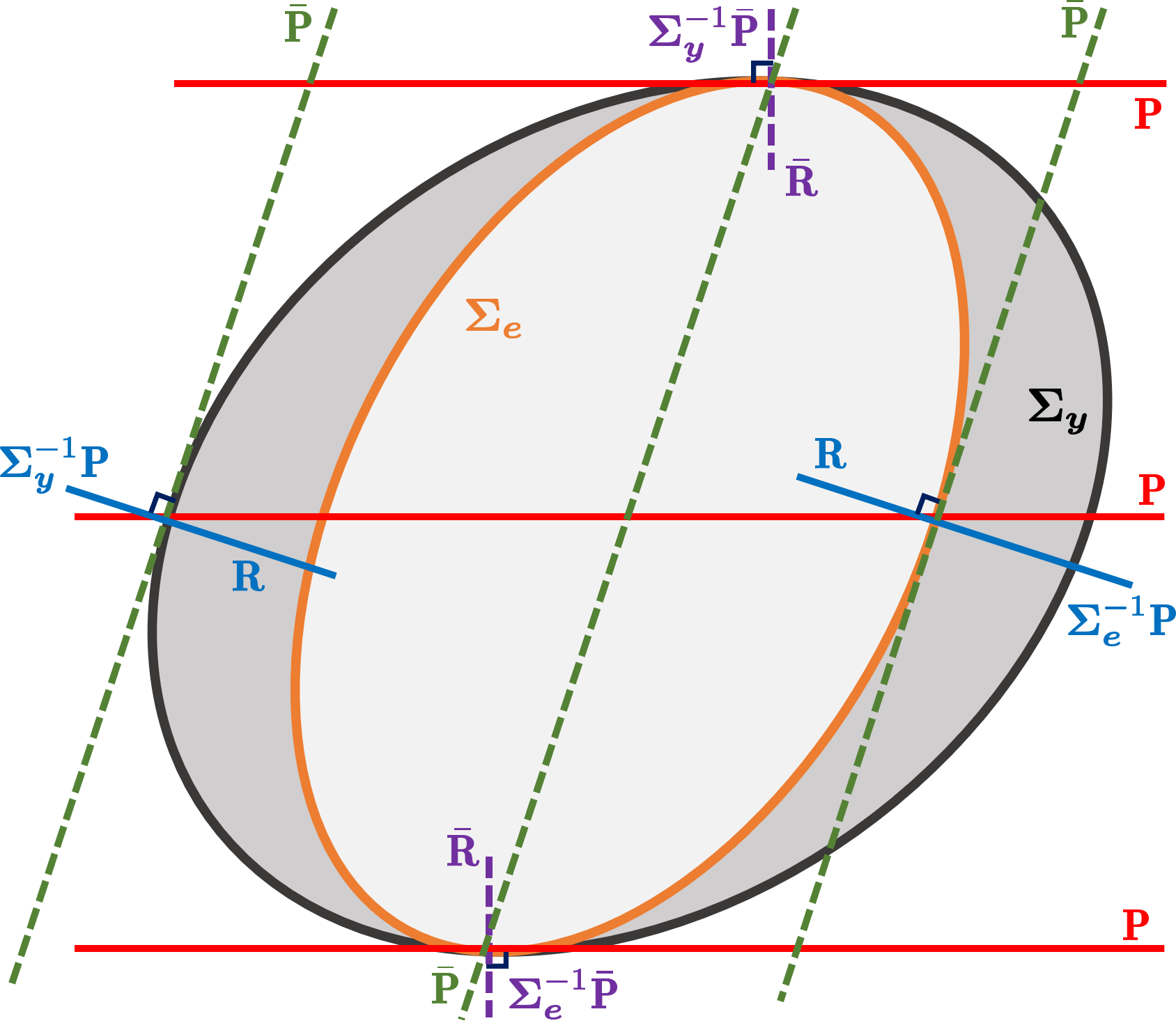}
  \caption{Oblique projection without normalization.}
  \label{fig:BeforeSTD}
\end{subfigure}%
\begin{subfigure}{.49\textwidth}
  \centering
  \includegraphics[height=0.6\textwidth]{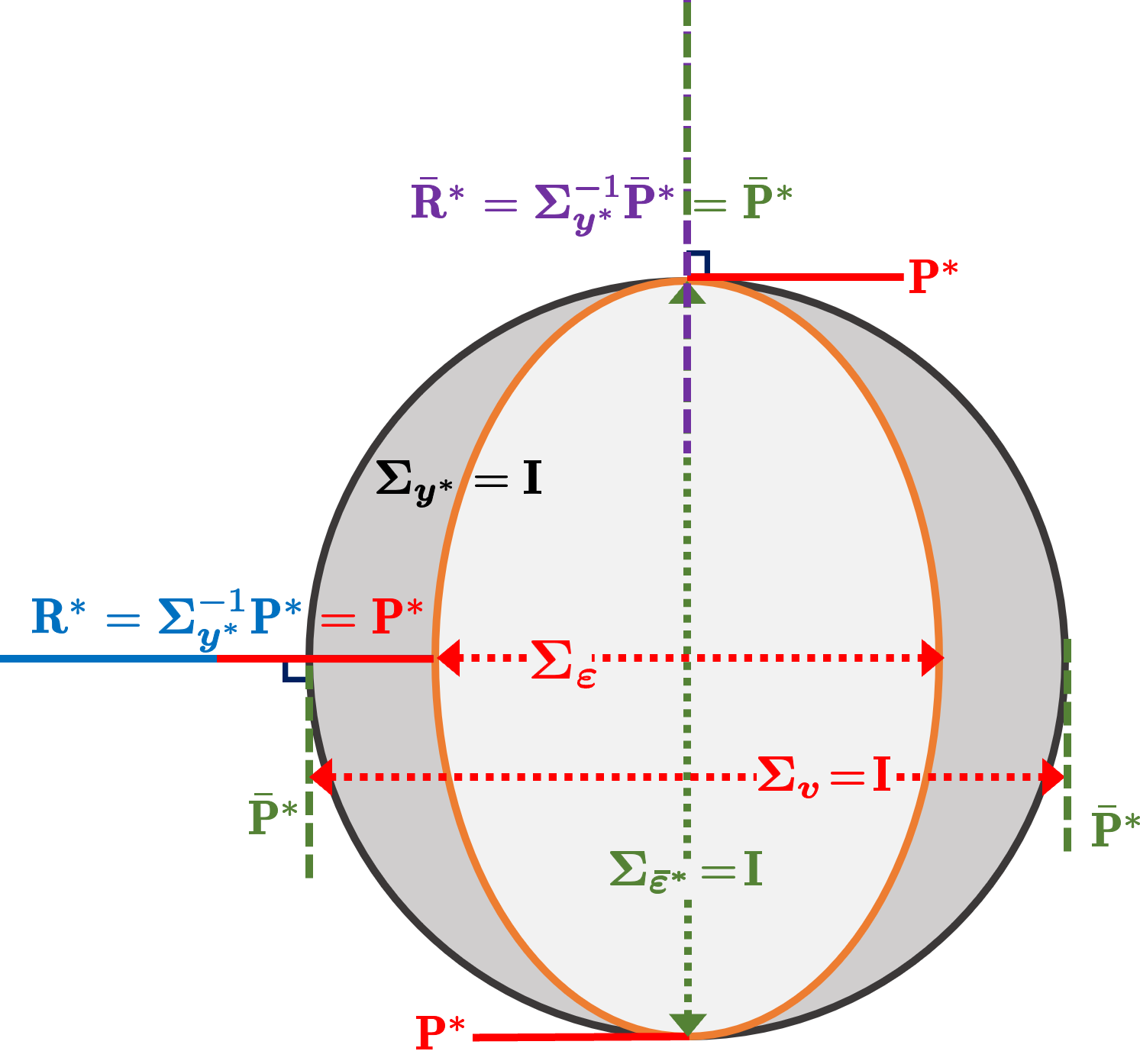}
  \caption{Normalization with~$\mathbf{\Sigma}_{\bm y^*}=\mathbf{I}$, $\mathbf{\Sigma}_{\bm v}=\mathbf{I}$, and~${\mathbf{R}^*}^\intercal{\mathbf{R}^*}=\mathbf{I}$.}
  \label{fig:AfterSTD}
\end{subfigure}
\caption{A geometric interpretation of optimal dynamic-static decomposition}
\label{fig:ellipsoid}
\end{figure*}

The PredVAR solution finds $\mathbf{R}$ to minimize the prediction error covariance of DLVs subject to $\mathbf{R}^\intercal\mathbf{P}=\mathbf{I}$. Thus, given~$\mathbf{P}$ and~$\mathbf{\Sigma}_{\bm e}$, we  solve the following optimization problem under the Loewner order~\cite{boyd2004convex}
\begin{align}~\label{eq:Sigma_varepsilon}
    \underset{\mathbf{R}}{\min\nolimits_\prec}~ \mathbf{\Sigma}_{\bm{\varepsilon}} = \mathbf{R}^\intercal\mathbf{\Sigma}_{\bm e}\mathbf{R} ~~~~\text{subject to}~~\mathbf{R}^\intercal\mathbf{P}=\mathbf{I},
\end{align}
for the best predictability. The optimization result leads to the following theorem that gives rise to uncorrelated innovations of $\bm{\varepsilon}_k$ and~$\bm{\bar \varepsilon}_k$.

\begin{theorem}\label{thm:eq_constraint}
When Problem~\eqref{eq:Sigma_varepsilon} attains the optimum,  the following statements hold and are equivalent.
\begin{enumerate}
    \item $ \mathbf{\Sigma}_{\bm e}\mathbf{R}=\mathbf{P}{\mathbf{R}}^\intercal\mathbf{\Sigma}_{\bm e}\mathbf{R}$;
    \item $\mathbf{P}{\mathbf{R}}^\intercal\bm{e}_k$ and~$(\mathbf{I}-\mathbf{P}{\mathbf{R}}^\intercal)\bm{e}_k$ are uncorrelated;
    \item $\bm{\varepsilon}_k$ and~$\bm{\bar \varepsilon}_k$ are uncorrelated;
    \item ${\mathbf{R}}^\intercal\mathbf{\Sigma}_{\bm e}\mathbf{\bar R}=\bm 0$; \label{eqst:REbarR}
    \item ${\mathbf{P}}^\intercal\mathbf{\Sigma}_{\bm e}^{-1}\mathbf{\bar P}=\bm 0$ if~$\mathbf{\Sigma}_{\bm e}$ is nonsingular; \label{eqst:PEbarP}
     \item ${\mathbf{R}}^\intercal \mathbf{\Sigma}_{\bm y}\mathbf{\bar R}=\bm 0$; \label{eqst:RYbarR}
     \item ${\mathbf{P}}^\intercal\mathbf{\Sigma}_{\bm y}^{-1}\mathbf{\bar P}=\bm 0$ if~$\mathbf{\Sigma}_{\bm y}$ is nonsingular. \label{eqst:PYbarP}
\end{enumerate}

\end{theorem}

Theorem \ref{thm:eq_constraint} is proven in Appendix~\ref{apd:eq_constraint}. The theorem reveals that minimizing the covariance of the DLV innovations leads to the uncorrelated realization as stated in the theorem. Thus, Algorithm \ref{alg:PredVAR} can be applied whenever a minimum covariance realization is desired.  The uncorrelated realization is a consequence of the oblique projection stated in \eqref{eq:Sigma_varepsilon}  that minimizes $\mathbf{\Sigma}_{\bm{\varepsilon}}$.

\subsection{Geometric Interpretation} \label{sec:geometric}

We illustrate the oblique projection necessitated by the covariance of the measurement vector $\bm y_k$ in Figs.~\ref{fig:BeforeSTD} and \ref{fig:AfterSTD}.
Fig.~\ref{fig:BeforeSTD} gives the oblique geometry of the optimal dynamic-static decomposition for the case of non-singular
$\mathbf{\Sigma}_{\bm y}$ and $\mathbf{\Sigma}_{\bm e}$, which are respectively represented by the ellipsoids
\[
\left\{\bm{x}\mid \bm{x}^\intercal\mathbf{\Sigma}_{\bm{y}}^{-1}\bm{x}\leq 1\right\} \text{ and } \left\{\bm{x}\mid \bm{x}^\intercal\mathbf{\Sigma}_{\bm{e}}^{-1}\bm{x}\leq 1\right\}.
\]
The two opposite tangent points of the ellipses indicate the static variance direction, which is irreducible by the RRVAR model prediction.
The tangent spaces of the two ellipsoids at their intersections on surfaces are the same, agreeing with the range of~$\mathbf{P}$. For the best predictability of DLVs in terms of~$\mathbf{\Sigma}_{\bm \varepsilon}$, a complementary subspace is chosen such that the oblique projection of the $\mathbf{\Sigma}_{\bm{e}}$-ellipsoid onto the dynamic subspace along the complementary subspace is as small as possible. Then, the complementary subspace should be specified by the tangent space to the $\mathbf{\Sigma}_{\bm{e}}$-ellipsoid at the intersection of the dynamic subspace passing the center and the surface. The corresponding normal space is the range of~$\mathbf{\Sigma}_{\bm e}^{-1}\mathbf{P}$, as shown in Fig.~\ref{fig:BeforeSTD}. Both $\mathbf{\Sigma}_{\bm e}^{-1}\mathbf{P}$ and $\mathbf{R}$ are orthogonal to $\mathbf{\bar P}$, aligning with Statement~\eqref{eqst:PEbarP} in Theorem~\ref{thm:eq_constraint}. Similarly, both $\mathbf{\Sigma}_{\bm e}\mathbf{\bar R}$ and $\mathbf{\bar P}$ are orthogonal to~$\mathbf{R}$, agreeing with Statement~\eqref{eqst:REbarR} in Theorem~\ref{thm:eq_constraint}.

On the other hand,
Fig.~\ref{fig:AfterSTD} illustrates the orthogonal geometry after normalization  with~$\mathbf{\Sigma}_{\bm y^*}=\mathbf{I}$. In this case, the optimal projection is normalized to an orthogonal one, making~$\mathbf{R}^*$ coincide with~$\mathbf{P}^*$. The covariances for the DLVs and static noise, $\mathbf{\Sigma}_{\bm v}$ and~$\mathbf{\Sigma}_{\bm {\bar \varepsilon} ^*}$, are scaled to identity matrices. Correspondingly, the predictability of the DLVs is visualized by the volume ratio of the two ellipsoids.

\section{Additional Analysis } \label{sec:add_analysis}

\subsection{Distributions of Parameter Estimates}
As in~\cite{reinsel:velu:1998}, we derive the statistical properties of the model estimates
under the assumption that the matrix~$\mathbb{V}$ of DLVs is fixed. First, it follows that
\begin{align*}
    E(\mathbb{\hat B}\mid \mathbb{V})&= E( ({\mathbb{V}}^{\intercal}{\mathbb{V}})^{-1}{\mathbb{V}}^{\intercal}\mathbf{ V}_{s}\mid \mathbb{V}) \\
    &= E(({\mathbb{V}}^{\intercal}{\mathbb{V}})^{-1}{\mathbb{V}}^{\intercal}(\mathbb{V}\mathbb{B}+\mathbf{E}^{\bm \varepsilon}_s)\mid \mathbb{V})=\mathbb{B},
\end{align*}
and
\begin{align*}
    &\textrm{Var}(\textrm{vec}(\mathbb{\hat B}^\intercal)\mid \mathbb{V})= \textrm{Var}((({\mathbb{V}}^{\intercal}{\mathbb{V}})^{-1}{\mathbb{V}}^{\intercal}\otimes \mathbf{I})\textrm{vec}(\mathbf{V}_s^\intercal)\mid \mathbb{V})\\
    =&\textrm{Var}((({\mathbb{V}}^{\intercal}{\mathbb{V}})^{-1}{\mathbb{V}}^{\intercal}\otimes \mathbf{I})\textrm{vec}(\mathbf{E}^{\bm \varepsilon^\intercal}_s))\mid \mathbb{V})\\
    =&E((({\mathbb{V}}^{\intercal}{\mathbb{V}})^{-1}{\mathbb{V}}^{\intercal}\!\otimes \mathbf{I})\! \times \!(\mathbf{I}\otimes \! \mathbf{\Sigma}_{\bm \varepsilon})\!\times \! (({\mathbb{V}}^{\intercal}{\mathbb{V}})^{-1}{\mathbb{V}}^{\intercal}\!\otimes\! \mathbf{I})^\intercal\mid \mathbb{V})\\
    =&({\mathbb{V}}^{\intercal}{\mathbb{V}})^{-1} \otimes \mathbf{\Sigma}_{\bm \varepsilon}.
\end{align*}
Therefore, $\mathbb{\hat B}$ is an unbiased estimate and $\textrm{Var}(\textrm{vec}(\mathbb{\hat B}^\intercal))$ can be assessed by replacing
$\mathbf{\Sigma}_{\bm \varepsilon}$ with its estimate.

 Similarly, since~$\mathbf{E}^{\bm e}_s=\mathbf{Y}_s-\mathbf{\hat V}_s\mathbf{P}^{\intercal}$ is uncorrelated to~$\mathbf{\hat V}_s$, it follows from~\eqref{update_P} that
\begin{align*}
    E(\mathbf{\hat P}\mid \mathbb{V})&= E( (\mathbf{E}^{\bm e}_s+\mathbf{\hat V}_s\mathbf{P}^{\intercal})^\intercal\mathbf{\hat V}_s (\mathbf{\hat V}_s^{\intercal} \mathbf{\hat V}_s)^{-1}\mid \mathbb{V}) =\mathbf{P},
    \end{align*}
and
    \begin{align*}
    \textrm{Var}(\textrm{vec}(\mathbf{\hat P})\mid \mathbb{V})&=E((\mathbf{\hat V}_s^{\intercal} \mathbf{\hat V}_s)^{-1}\otimes \mathbf{\Sigma}_{\bm e}\mid \mathbb{V})\\
    &=(\mathbb{B}^{\intercal}\mathbb{V}^{\intercal} \mathbb{V}\mathbb{B})^{-1}\otimes \mathbf{\Sigma}_{\bm e}.
\end{align*}
Again, $\mathbf{\Sigma}_{\bm e}$ can be replaced by its estimate to assess $\textrm{Var}(\textrm{vec}(\mathbf{\hat P}))$.
To sum up, we conclude that the parameter estimates can be approximated with the following distributions given  $\mathbb{V}$:
\begin{align*}
   & \textrm{vec}(\mathbb{\hat B^\intercal})\sim {\mathcal N} (\textrm{vec}(\mathbb{B}^\intercal), ({\mathbb{V}}^{\intercal}{\mathbb{V}})^{-1}\otimes \mathbf{\hat  \Sigma}_{\bm \varepsilon})\text{ and } \\
& \textrm{vec}(\mathbf{\hat P}) \sim
    {\mathcal N} (\textrm{vec}(\mathbf{ P}),(\mathbb{B}^{\intercal}\mathbb{V}^{\intercal} \mathbb{V}\mathbb{B})^{-1}\otimes \mathbf{\hat \Sigma}_{\bm e}).
\end{align*}

\subsection{Selection of Model Sizes}\label{sec:FPE}


For full-rank VAR models, the multiple final prediction error (MFPE) developed by  Akaike~\cite{akaike1971autoregressive} can be used to determine the  order~$s$. Following this idea, we derive {a reduced-rank  MFPE (RRMFPE)} to determine~$s$ and~$\ell$ for the RRVAR model. We choose to work on the normalized vector~$\bm y_k^*$ since  $\mathbf{P}^*$ and $\mathbf{R}^*$ coincide with each other and  $\mathbf{P}^*$ and  $\mathbf{\bar P}^*$ are orthogonal. Therefore, we can write the model explicitly conditional on $\mathbf{\hat P}^*$ as follows
 based on \eqref{eq:PredVAR_yk*}
 \begin{align*}
\bm{y}_{k|\mathbf{\hat P}^*}^*
    =\begin{bmatrix}
   \mathbf{\hat P}^*&   \mathbf{\hat{\bar P}^*}
    \end{bmatrix}
\begin{bmatrix}
    \sum_{j=1}^s \mathbf{B}_{j|\mathbf{\hat P}^*} \bm{v}_{k-j|\mathbf{\hat P}^*}+\bm{\varepsilon}_{k|\mathbf{\hat P}^*}\\
    \bm{\bar \varepsilon}_{k|\mathbf{\hat{ P}^*}}^*
\end{bmatrix}
\end{align*}
and the prediction error based on the estimates~$\{{\mathbf{\hat B}_{j|\mathbf{\hat P}^*}}\}$ is given as follows
\begin{align*}
    &\bm{\tilde y}_{k|\mathbf{\hat P}^*}^* = \bm{y}_{k|\mathbf{\hat P}^*}^* -\bm{\hat y}_{k|\mathbf{\hat P}^*}^*\\
    =&\begin{bmatrix}
   \mathbf{\hat P}^*&
    \mathbf{\hat{\bar P}^*}
    \end{bmatrix}\begin{bmatrix}
    \sum_{j=1}^s(\mathbf{B}_{j|\mathbf{\hat P}^*}-\mathbf{\hat B}_{j|\mathbf{\hat P}^*})\bm{v}_{k-j|\mathbf{\hat P}^*}+\bm{\varepsilon}_{k|\mathbf{\hat P}^*}\\
    \bm{\bar \varepsilon}_{k|\mathbf{\hat{ P}^*}}^*
\end{bmatrix}.
\end{align*}

The covariance of the prediction error is
\begin{multline*}
    E\left( \bm{\tilde y}_{k|\mathbf{\hat P}^*}^*
    \bm{\tilde y}_{k|\mathbf{\hat P}^* }^{*\intercal} \right)=\begin{bmatrix}
   \mathbf{\hat P}^*&
    \mathbf{\hat{\bar P}^*}
    \end{bmatrix}\\\begin{bmatrix}
    \mathbf{\hat \Sigma}_{\bm \varepsilon}+\!\!\!\sum\limits_{i,j=1}^s\!\!\! \mathit{\Delta}\mathbf{B}_{i|\mathbf{\hat P}^*}\bm{v}_{k-i|\mathbf{\hat P}^*}\bm{v}_{k-j|\mathbf{\hat P}^*}^\intercal\!\mathit{\Delta}\mathbf{B}_{j|\mathbf{\hat P}^*}^\intercal& \mathbf{\hat \Sigma}_{\bm{\varepsilon}, \bm{\bar \varepsilon}^*}\\
    \mathbf{\hat \Sigma}_{\bm{\bar \varepsilon}^*, \bm{\varepsilon}} &\mathbf{\hat \Sigma}_{\bm{\bar \varepsilon}^*}
\end{bmatrix}\begin{bmatrix}
   \mathbf{\hat P}^{*\intercal}\\
    \mathbf{\hat{\bar P}^{*\intercal}}
    \end{bmatrix},
\end{multline*}
where~$\mathit{\Delta}\mathbf{B}_{i|\mathbf{\hat P}^*}=\mathbf{B}_{i|\mathbf{\hat P}^*}-\mathbf{\hat B}_{i|\mathbf{\hat P}^*}$ and $\mathbf{\hat \Sigma}_{\bm{\varepsilon},\bm{\bar \varepsilon}^*} = \mathbf{0} $
for the PredVAR algorithm. Following \cite{akaike1971autoregressive} and using the fact that~$\mathbf{\hat B}_{j|\mathbf{\hat P}^*}$ from~\eqref{update_B} is a least-square estimate with respect to~$\{\bm{v}_{k|\mathbf{\hat P}^*}\}$, the estimate of the above covariance is
\begin{multline*}
    {\hat E}\left( \bm{\tilde y}_{k|\mathbf{\hat P}^*}^*      \bm{\tilde y}_{k|\mathbf{\hat P}^* }^{*\intercal}  \right)=
    \begin{bmatrix}
   \mathbf{\hat P}^*&
    \mathbf{\hat{\bar P}^*}
    \end{bmatrix}\begin{bmatrix}
    \frac{1+ s  \ell/N}{1- s  \ell/N} \mathbf{\hat \Sigma}_{\bm \varepsilon} & \mathbf{0} \\
    \mathbf{0} &\mathbf{\hat \Sigma}_{\bm{\bar \varepsilon}^*}\end{bmatrix}\begin{bmatrix}
   \mathbf{\hat P}^{*\intercal}\\
    \mathbf{\hat{\bar P}^{*\intercal}}
    \end{bmatrix}.
\end{multline*}
Since~$\begin{bmatrix}
   \mathbf{\hat P}^*&
    \mathbf{\hat{\bar P}^*}
    \end{bmatrix}^\intercal\begin{bmatrix}
   \mathbf{\hat P}^*&
    \mathbf{\hat{\bar P}^*}
    \end{bmatrix} = \mathbf{I}$, taking the determinant of the above covariance estimate gives the {RRMFPE} of the PredVAR algorithm with the model size~$(\ell,s)$ as follows
 \begin{align}
 \texttt{RRMFPE}(\ell,s) &= \textrm{det}\begin{bmatrix}
 \frac{1+ s  \ell/N}{1- s  \ell/N} \mathbf{\hat \Sigma}_{\bm \varepsilon} & \mathbf{0}\\
    \mathbf{0} &\mathbf{\hat \Sigma}_{\bm{\bar \varepsilon}^*}
\end{bmatrix} \nonumber\\
&=
    \left( \frac{1+ s  \ell/N}{1- s  \ell/N} \right)^\ell \textrm{det}(\mathbf{\hat \Sigma}_{  {\bm \varepsilon}}) \textrm{det}(\mathbf{\hat \Sigma}_{\bm{\bar \varepsilon}^*}) \nonumber\\
&=
    \left( \frac{1+ s  \ell/N}{1- s  \ell/N} \right)^\ell \Pi_{i=1}^\ell (1 - \lambda_i), \label{eq:RRMFPE}
 \end{align}
using
$\mathbf{\hat \Sigma}_{\bm{\bar \varepsilon}^*} = \mathbf{I}$ and
$\textrm{det}(\mathbf{\hat \Sigma}_{  {\bm \varepsilon}})
= \Pi_{i=1}^\ell (1 - \lambda_i)$ from \eqref{eq:Sigma_epsilon}. To handle the possible rank-deficient
$\mathbf{\hat \Sigma}_{  {\bm \varepsilon}}$ due to perfect prediction of some DLVs, denote~{${\mathcal S}_+ = \{i \mid 1 - \lambda_i>0, \text{ for }i = 1, \cdots, \ell\}$ and exclude the zero variance terms by replacing
$\textrm{det}(\mathbf{\hat \Sigma}_{  {\bm \varepsilon}})$ with
$ \Pi_{i\in {\mathcal S}_+} (1 - \lambda_i)$. To further simplify the expression, we take logarithm to obtain
 \begin{align*}
 \texttt{log-RRMFPE}(\ell,s) =
  \ell \log   \frac{1+ s  \ell/N}{1- s  \ell/N} + \sum_{i\in {\mathcal S}_+} \log  (1 - \lambda_i).
 \end{align*}
When~$N$ is large enough, taking the first-order Taylor approximation gives
 \begin{align} \label{eq:log-RRMFPE}
 \texttt{log-RRMFPE}(\ell,s) =
  \sum_{i\in {\mathcal S}_+} \log  (1 - \lambda_i)
  + 2 \frac{s \ell^2 }{N}.
 \end{align}
We can fit the model with a grid of~$(\ell, s)$-pairs, and the one that gives the minimum log-RRMFPE can be chosen as the optimal model size. The two terms in \eqref{eq:log-RRMFPE} give a clear trade-off between model complicity and model errors. With fixed $\ell$, the first term decreases with $s$, while the second term increases linearly with $s$. On the other hand, with fixed $s$, the first term decreases with $\ell$, while the second term increases quadratically with $\ell$.


\subsection{Comparison among~\modelname, LaVAR-CCA, and DiCCA Algorithms}

Since the PredVAR, LaVAR-CCA \cite{Qin:2022LaVAR_AIChEJ}, and  DiCCA \cite{dong:liu:qin:2020DICCA-SVD} algorithms all use equivalent normalization of the data, it is convenient to compare them using the normalized data~$\{\bm y^*_k\}$. LaVAR-CCA~\cite{Qin:2022LaVAR_AIChEJ} essentially updates~$\mathbf{\hat R}^*$ by solving the following optimization problem
\begin{equation*}
        \underset{\mathbf{\hat R}^{*\intercal} \mathbf{\hat R}^*=\mathbf{I}}{\max \nolimits_\prec}~~ \mathbf{\hat R}^{*\intercal}\left(\mathbf{Y}_s^{*\intercal} \mathbf{\Pi}_{\mathbb{ V}} \mathbf{Y}_s^*/N \right)\mathbf{\hat R}^* 
\end{equation*}
Comparing to \eqref{eq:Sigma_v hat},
 the proposed PredVAR algorithm updates~$\mathbf{\hat R}^*$ by
 \begin{equation*}
        \underset{\mathbf{\hat R}^{*\intercal} \mathbf{\hat R}^*=\mathbf{I}}{\max \nolimits_\prec}~~ \mathbf{\hat R}^{*\intercal}\left(\mathbf{Y}_s^{*\intercal} \mathbf{\Pi}_{\mathbf{\hat V}_s} \mathbf{Y}_s^*/N \right)\mathbf{\hat R}^* 
\end{equation*}
Therefore, the two algorithms are different in general.
 As per~\eqref{update_Vs_hat}, the range of~$\mathbf{\hat V}_s$ is contained in that of~$\mathbb{ V}$, that is,  $ Span (\mathbf{\hat V}_s) \subseteq Span (\mathbb{V})$. A projection onto the former implies a more parsimonious model. Consequently, the PredVAR algorithm focuses on the portion of the past information that is relevant to predicting the current data. Therefore, the PredVAR algorithm makes better use of the latent dynamics for the optimal dimension reduction than LaVAR-CCA. The two algorithms coincide in the special case of $s=1$ since the range of~$\mathbf{\hat V}_s$ is the same as that of~$\mathbb{ V}$.

The difference between \modelname~and DiCCA is evident since DiCCA works on univariate latent dynamics. For the special case of $\ell=1$, DiCCA and LaVAR-CCA use the same objective. However, the difference between  \modelname~and DiCCA remains since the range of~$\mathbf{\hat V}_s$ is contained in that of~$\mathbb{ V}$.




\subsection{Comparison to Non-iterative One-Shot Solutions }\label{Sec:OS}

The PredVAR, LaVAR-CCA, and DiCCA algorithms are all iterative to find~$\mathbf{\hat R}^*$. In the literature, papers like~\cite{gao:tsay:2021high-Dim} adopt non-iterative one-shot (OS) solutions, which do not enforce an explicit latent VAR model to obtain~$\mathbf{\hat R}^*$.  The DLV sequences are calculated and used to retrofit the VAR model matrices similar to~\eqref{update_B}. However, the estimated VAR model is not used to further update~$\mathbf{\hat R}^*$. Thus, OS solutions appear to be a single iteration step of the iterative algorithms like the PredVAR.

We consider the OS algorithm developed in~\cite{gao:tsay:2021high-Dim} and compare it with PredVAR. The work in~\cite{gao:tsay:2021high-Dim} first normalizes the data and obtains an orthonormal~$\mathbf{\hat{\bar R}^*}$ from the eigenvectors corresponding to the least~$p-\ell$ eigenvalues of
\begin{align} \label{eq:EVD_Gao-Tsay}
  \sum_{k=1}^{s} (\mathbf{Y}_s^{*\intercal}\mathbf{Y}_{s-k}^*)(\mathbf{Y}_s^{*\intercal}\mathbf{Y}_{s-k}^*)^{\intercal}
= \mathbf{Y}_s^{*\intercal} \mathbb{Y}^* \mathbb{Y}^{*\intercal} \mathbf{Y}_s^*,
\end{align}
where~$s$ is the same as $k_0$  in~\cite{gao:tsay:2021high-Dim} as a pre-specified integer. Using
$\mathbf{\hat{\bar R}^*}$  to obtain $\mathbf{\hat{\bar R}}$,
$\mathbf{\hat R}$ is found to be the eigenvectors associated with the $\ell$ smallest eigenvalues of $\mathbf{\hat \Sigma}_{\bm y}\mathbf{\hat{\bar R}}\mathbf{\hat{\bar R}}^{\intercal} \mathbf{\hat \Sigma}_{\bm y}$. In other words,
$\mathbf{\hat R}^\intercal \mathbf{\hat \Sigma}_{\bm y}\mathbf{\hat{\bar R}} \mathbf{\hat{\bar R}}^{\intercal} \mathbf{\hat \Sigma}_{\bm y} \mathbf{\hat R}$ is a diagonal matrix of the $\ell$ smallest eigenvalues. This solution is consistent with \eqref{eqst:RYbarR} in Theorem \ref{thm:eq_constraint}, which makes
$\mathbf{\hat R}^\intercal \mathbf{\hat \Sigma}_{\bm y}\mathbf{\hat{\bar R}} $ zero ideally.

It is apparent that the above OS solution resembles the initialization step of PredVAR in \eqref{eq:EVD_initialize}, but ignores the subsequent iterations of the PredVAR algorithm. The OS solution from \eqref{eq:EVD_Gao-Tsay} finds the left singular vectors of
$\mathbf{Y}_s^{*\intercal} \mathbb{Y}^*$, which is essentially the covariance from the two matrices. On the other hand, the initialization step of PredVAR in \eqref{eq:EVD_initialize} makes use of the eigenvectors of
$\mathbf{Y}_s^{*\intercal} \mathbb{Y}^*
(\mathbb{Y}^{*\intercal} \mathbb{Y}^{*})^{-1} \mathbb{Y}^{*\intercal} \mathbf{Y}_s^*$, which are projections of $\mathbf{Y}_s^*$ onto
$Span (\mathbb{Y}^{*})$, or the left singular vectors of
$\mathbf{Y}_s^{*\intercal} \mathbb{Y}^*
(\mathbb{Y}^{*\intercal} \mathbb{Y}^{*})^{-\frac{1}{2}}$. Since the covariance of $\mathbf{Y}_s^*$ is identity, $\mathbf{Y}_s^{*\intercal} \mathbb{Y}^*
(\mathbb{Y}^{*\intercal} \mathbb{Y}^{*})^{-\frac{1}{2}}$ contains the canonical correlations from the two matrices.

 To summarize, one-shot solutions like \cite{gao:tsay:2021high-Dim} resemble the initialization step of iterative solutions such as PredVAR and do not iterate further to enforce consistency of the outer projection and inner dynamic VAR models. In addition, \cite{gao:tsay:2021high-Dim}  relies on the covariance
$\mathbf{Y}_s^{*\intercal} \mathbb{Y}^*$ instead of the correlations
$\mathbf{Y}_s^{*\intercal} \mathbb{Y}^*
(\mathbb{Y}^{*\intercal} \mathbb{Y}^{*})^{-\frac{1}{2}}$ to find the solution. The difference between the two is the weight matrix
$(\mathbb{Y}^{*\intercal} \mathbb{Y}^{*})^{-\frac{1}{2}}$, whose effect has been studied in the subspace identification literature extensively \cite{vanoverschee:demoor:1996,Ljung:1999,chiuso:picci:2005}. It is noted that the weight matrix does affect the singular vector solutions.

\section{Case Studies} \label{sec:simulation}





In this section, we conduct comprehensive numerical comparisons of the proposed PredVAR algorithm with other leading algorithms,  including the LaVAR-CCA (abbreviated as LaVAR) in~\cite{Qin:2022LaVAR_AIChEJ}, DiCCA in~\cite{dong:liu:qin:2020DICCA-SVD}, and the one-shot algorithm in~\cite{gao:tsay:2021high-Dim} (see Sec.~\ref{Sec:OS}). All simulations are conducted via MATLAB with an Apple M2 Pro.

\begin{table*}[ht!]
\caption{Performance of \modelname~under different model sizes~$(\ell, s)$.}
     \label{tab:DlDs}
    \begin{subtable}[h]{0.45\textwidth}
        \centering
        \scriptsize
        \setlength\tabcolsep{5pt}
        \renewcommand{\arraystretch}{1.1}
        \begin{tabularx}{1\linewidth}{c|cccccc}
  \toprule    \toprule
 \diagbox{$\ell$}{$s$} & 2 & 12 & 22 & 32 & 42 &52 \\\hline
1 & 0.8265 & 0.8198 & 0.8198 & 0.8198 & 0.8198 & 0.8198 \\
2 & 0.6032 & 0.5918 & 0.5918 & 0.5918 & 0.5918 & 0.5918 \\
3 & 0.1891 & 0.1895 & 0.1895 & 0.1895 & 0.1896 & 0.1895 \\
4 & 0.5142 & 0.5132 & 0.5129 & 0.5127 & 0.5126 & 0.5123 \\
5 & 0.6385 & 0.6379 & 0.6381 & 0.6386 & 0.6388 & 0.6383 \\
6 & 0.7071 & 0.7071 & 0.7071 & 0.7071 & 0.7071 & 0.7071 \\
  \bottomrule \bottomrule
  \end{tabularx}
       \caption{D-distance between the ranges of~$\mathbf{\hat P}$ and~$\mathbf{P}$.}
       \label{tab:DlDs_angleP}
    \end{subtable}
    \hfill
    \begin{subtable}[h]{0.45\textwidth}
        \centering
        \scriptsize
        \setlength\tabcolsep{5pt}
        \renewcommand{\arraystretch}{1.1}
        \begin{tabularx}{1\linewidth}{c|cccccc}
  \toprule    \toprule
 \diagbox{$\ell$}{$s$} & 2 & 12 & 22 & 32 & 42 &52 \\\hline
1 & 0.3129 & 0.7641 & 0.7599 & 0.7592 & 0.7589 & 0.7586 \\
2 & 0.6349 & 0.9166 & 0.9123 & 0.9119 & 0.9117 & 0.9117 \\
3 & 0.9995 & 0.9995 & 0.9995 & 0.9995 & 0.9995 & 0.9995 \\
4 & 0.7658 & 0.7543 & 0.7599 & 0.7714 & 0.7670 & 0.7716 \\
5 & 0.6995 & 0.7086 & 0.7002 & 0.7242 & 0.7489 & 0.6853 \\
6 & 0.6501 & 0.6501 & 0.6501 & 0.6501 & 0.6501 & 0.6501 \\
  \bottomrule \bottomrule
  \end{tabularx}
       \caption{Average correlation between~$\{\mathbf{\hat P}\bm{ v}_{k|\mathbf{\hat R}}\}$ and~$\{\mathbf{P}\bm{v}_{k}\}$.}
       \label{tab:DlDs_proj}
    \end{subtable}
    \hfill
\begin{subtable}[h]{0.45\textwidth}
        \centering
        \scriptsize
        \setlength\tabcolsep{5pt}
        \renewcommand{\arraystretch}{1.1}
        \begin{tabularx}{1\linewidth}{c|cccccc}
  \toprule    \toprule
 \diagbox{$\ell$}{$s$} & 2 & 12 & 22 & 32 & 42 &52\\\hline
1 & 0.3112 & 0.7641 & 0.7599 & 0.7590 & 0.7583 & 0.7573 \\
2 & 0.6332 & 0.9166 & 0.9123 & 0.9120 & 0.9120 & 0.9119 \\
3 & 0.9974 & 0.9698 & 0.9994 & 0.9994 & 0.9995 & 0.9995 \\
4 & 0.9987 & 0.9897 & 0.9870 & 0.9851 & 0.9827 & 0.9806 \\
5 & 0.9981 & 0.9869 & 0.9833 & 0.9769 & 0.9781 & 0.9688 \\
6 & 0.9977 & 0.9837 & 0.9759 & 0.9680 & 0.9616 & 0.9546 \\
  \bottomrule \bottomrule
  \end{tabularx}
       \caption{Average correlation between~$\{\mathbf{\hat P}\bm{\hat{{v}}}_{k|\mathbf{\hat R}}\}$ and~$\{\mathbf{P}\bm{v}_{k}\}$.}
       \label{tab:DlDs_pred}
    \end{subtable}
    \hfill
\begin{subtable}[h]{0.45\textwidth}
        \centering
        \scriptsize
        \setlength\tabcolsep{5pt}
        \renewcommand{\arraystretch}{1.1}
        \begin{tabularx}{1\linewidth}{c|cccccc}
  \toprule    \toprule
 \diagbox{$\ell$}{$s$} & 2 & 12 & 22 & 32 & 42 &52 \\\hline
1 & 0.9993 & 1.0000 & 1.0000 & 1.0000 & 1.0000 & 1.0000 \\
2 & 0.9997 & 1.0000 & 1.0000 & 1.0000 & 1.0000 & 1.0000 \\
3 & 0.9975 & 0.9680 & 0.9999 & 0.9999 & 0.9999 & 0.9999 \\
4 & 0.7657 & 0.7517 & 0.7561 & 0.7678 & 0.7621 & 0.7673 \\
5 & 0.6990 & 0.7037 & 0.6937 & 0.7199 & 0.7430 & 0.6718 \\
6 & 0.6495 & 0.6436 & 0.6409 & 0.6396 & 0.6348 & 0.6324 \\
  \bottomrule \bottomrule
  \end{tabularx}
       \caption{Average correlation between~$\{\mathbf{\hat P}\bm{ \hat v}_{k|\mathbf{\hat R}}\}$ and~$\{\mathbf{\hat P}\bm{ v}_{k|\mathbf{\hat R}}\}$.}
       \label{tab:DlDs_in}
    \end{subtable}
        \hfill
\begin{subtable}[h]{0.45\textwidth}
        \centering
        \scriptsize
        \setlength\tabcolsep{5pt}
        \renewcommand{\arraystretch}{1.1}
        \begin{tabularx}{1\linewidth}{c|cccccc}
  \toprule    \toprule
 \diagbox{$\ell$}{$s$} & 2 & 12 & 22 & 32 & 42 &52 \\\hline
1 & 0.1633 & 0.4891 & 0.4876 & 0.4873 & 0.4872 & 0.4870 \\
2 & 0.4127 & 0.6072 & 0.6056 & 0.6055 & 0.6054 & 0.6054 \\
3 & 0.6505 & 0.6505 & 0.6505 & 0.6505 & 0.6505 & 0.6505 \\
4 & 0.8634 & 0.8763 & 0.8689 & 0.8542 & 0.8583 & 0.8480 \\
5 & 0.9424 & 0.9231 & 0.9288 & 0.9094 & 0.8725 & 0.9439 \\
6 & 1.0000 & 1.0000 & 1.0000 & 1.0000 & 1.0000 & 1.0000 \\
  \bottomrule \bottomrule
  \end{tabularx}
       \caption{Average correlation between~$\{\mathbf{\hat P}\bm{  v}_{k|\mathbf{\hat R}}\}$ and~$\{\bm y_k\}$.}
       \label{tab:DlDs_out}
    \end{subtable}
    \hfill
\begin{subtable}[h]{0.45\textwidth}
        \centering
        \scriptsize
        \setlength\tabcolsep{5pt}
        \renewcommand{\arraystretch}{1.1}
        \begin{tabularx}{1\linewidth}{c|cccccc}
  \toprule    \toprule
 \diagbox{$\ell$}{$s$} & 2 & 12 & 22 & 32 & 42 &52 \\\hline
1 & 0.1623 & 0.4895 & 0.4883 & 0.4888 & 0.4881 & 0.4873 \\
2 & 0.4117 & 0.6076 & 0.6063 & 0.6070 & 0.6068 & 0.6066 \\
3 & 0.6491 & 0.6305 & 0.6512 & 0.6515 & 0.6514 & 0.6513 \\
4 & 0.6503 & 0.6470 & 0.6468 & 0.6474 & 0.6463 & 0.6461 \\
5 & 0.6497 & 0.6452 & 0.6437 & 0.6452 & 0.6446 & 0.6380 \\
6 & 0.6495 & 0.6436 & 0.6409 & 0.6396 & 0.6348 & 0.6324 \\
  \bottomrule \bottomrule
  \end{tabularx}
       \caption{Average correlation between~$\{\mathbf{\hat P}\bm{  \hat v}_{k|\mathbf{\hat R}}\}$ and~$\{\bm y_k\}$.}
       \label{tab:DlDs_inout}
    \end{subtable}
\end{table*}

\subsection{Case Study with the Lorenz Attractor Data}
As in~\cite{Qin:2022LaVAR_AIChEJ}, use the nonlinear Lorenz oscillator to generate~$10,000$  samples for the DLV~$\{\bm v_k \in \Re^3\}$. The static noise~$\{\bm{\bar \varepsilon}_k\in\Re^3\}$ is generated following a zero-mean Gaussian distribution whose variance is the same as that of the DLV series. Then, the~$6$-dimensional measurement samples~$\{\bm y_k\}$ is obtained by mixing the DLVs and static noise via~\eqref{eq:outer_model}, where~$\mathbf{P},\mathbf{\bar P}\in \Re^{6\times 3}$ are orthonormal and~$[\mathbf{P} ~~ \mathbf{\bar P}]$ is invertible. The first $7,000$ measurement samples are for training, and the remaining is for testing.


\begin{figure*}[ht!]
     \centering
     \begin{subfigure}[t]{0.49\textwidth}
         \centering
         \includegraphics[width=0.9\textwidth]{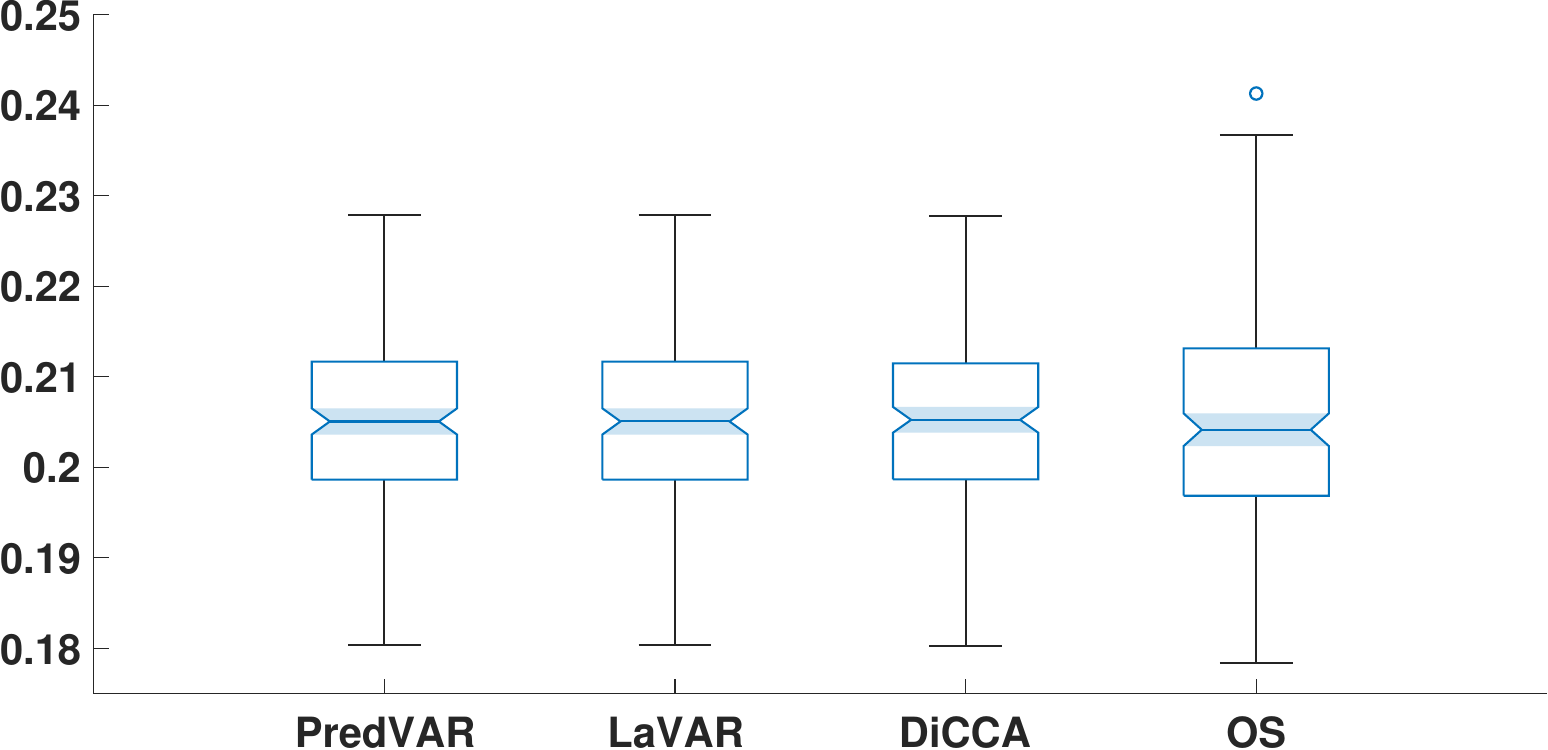}
         \caption{D-distance between the ranges of~$\mathbf{\hat P}$ and $\mathbf{P}$.}
         \label{fig:Rep_angleP}
     \end{subfigure}
     \hfill
     \begin{subfigure}[t]{0.49\textwidth}
         \centering
         \includegraphics[width=0.9\textwidth]{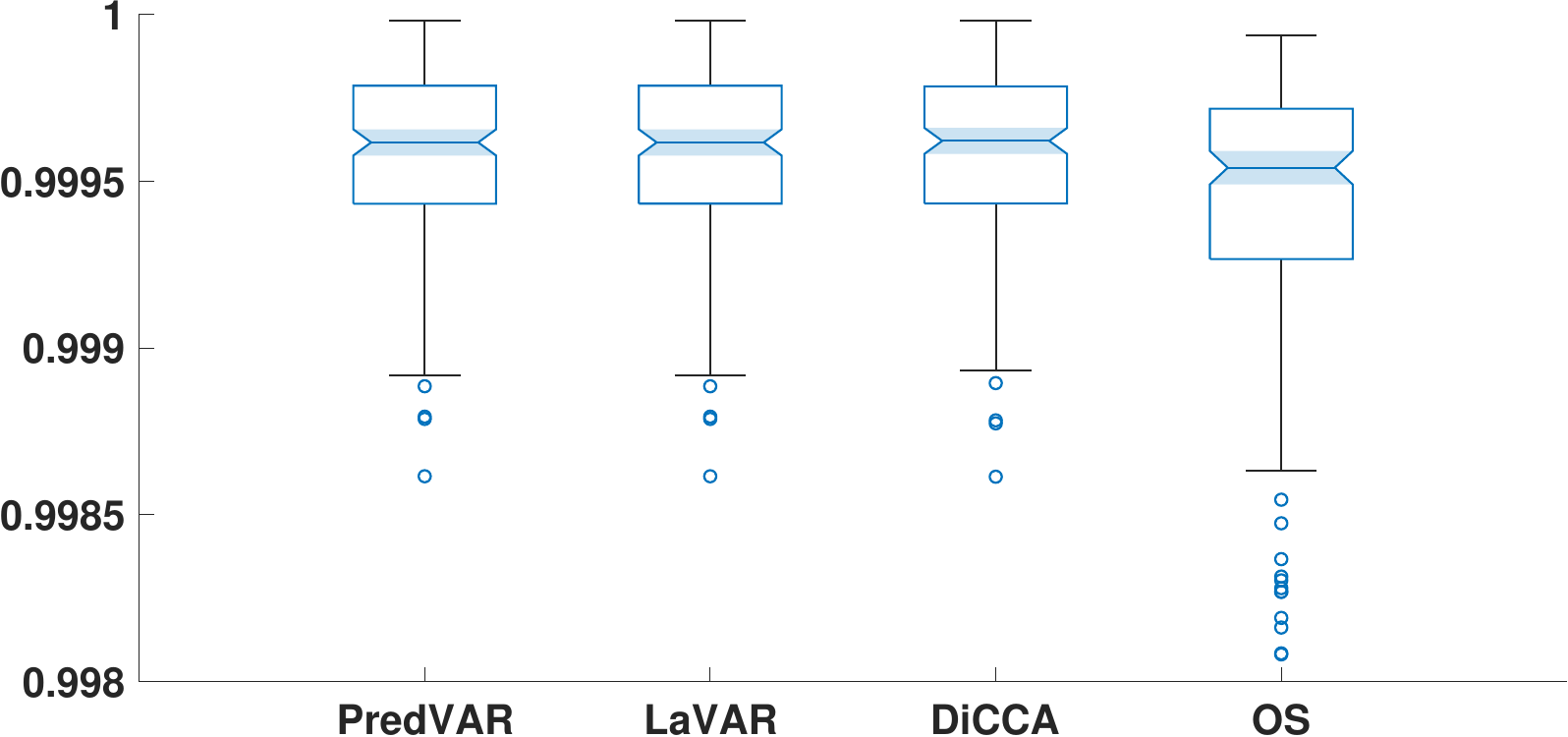}
         \caption{Average correlation between~$\{\mathbf{\hat P}\bm{ v}_{k|\mathbf{\hat R}}\}$ and~$\{\mathbf{P}\bm{v}_{k}\}$.}
         \label{fig:Rep_proj}
     \end{subfigure}
     \hfill
     \begin{subfigure}[t]{0.49\textwidth}
         \centering
         \includegraphics[width=0.9\textwidth]{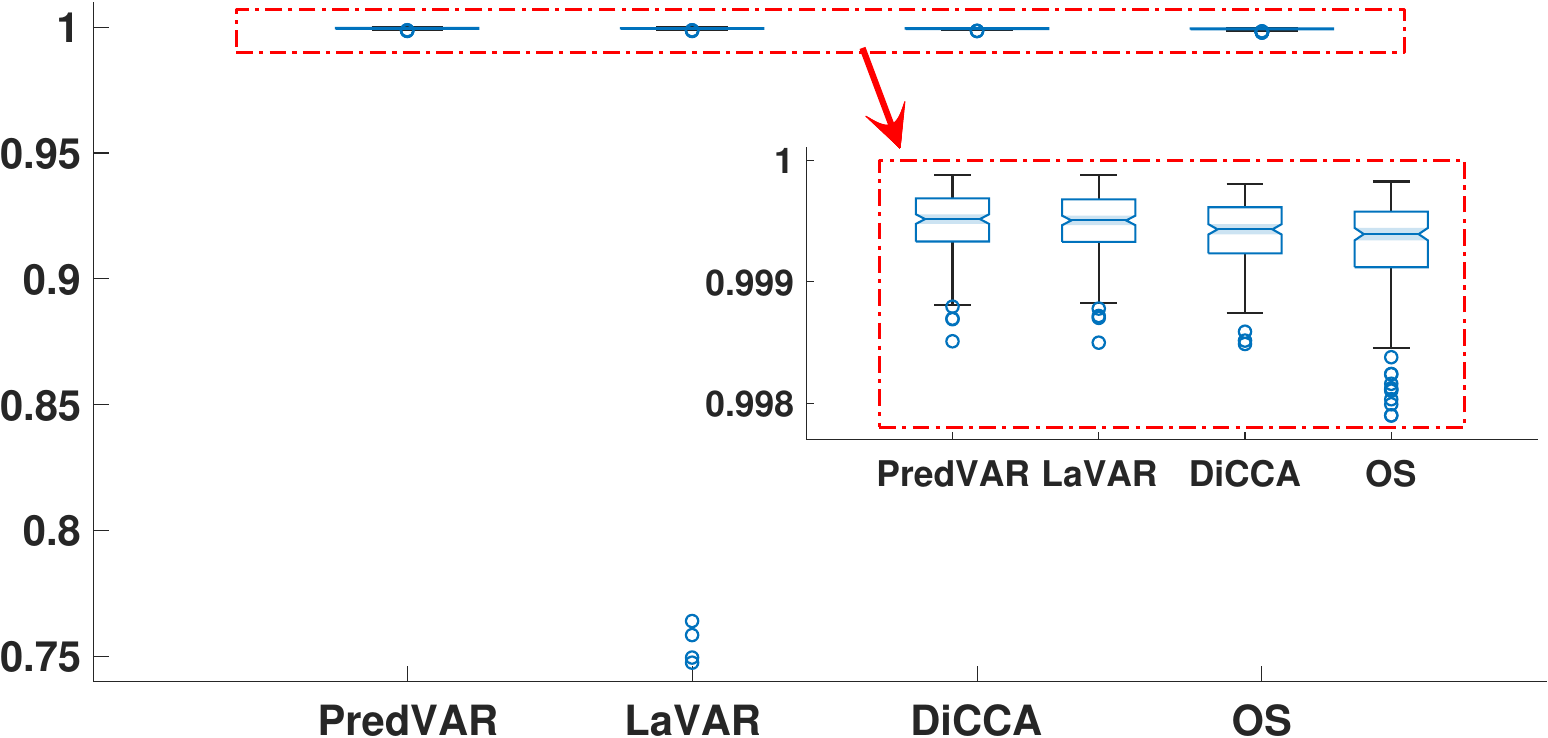}
         \caption{Average correlation between~$\{\mathbf{\hat P}\bm{ \hat v}_{k|\mathbf{\hat R}}\}$ and~$\{\mathbf{P}\bm{v}_{k}\}$.}
         \label{fig:Rep_pred}
     \end{subfigure}
     \hfill
     \begin{subfigure}[t]{0.49\textwidth}
         \centering
         \includegraphics[width=0.9\textwidth]{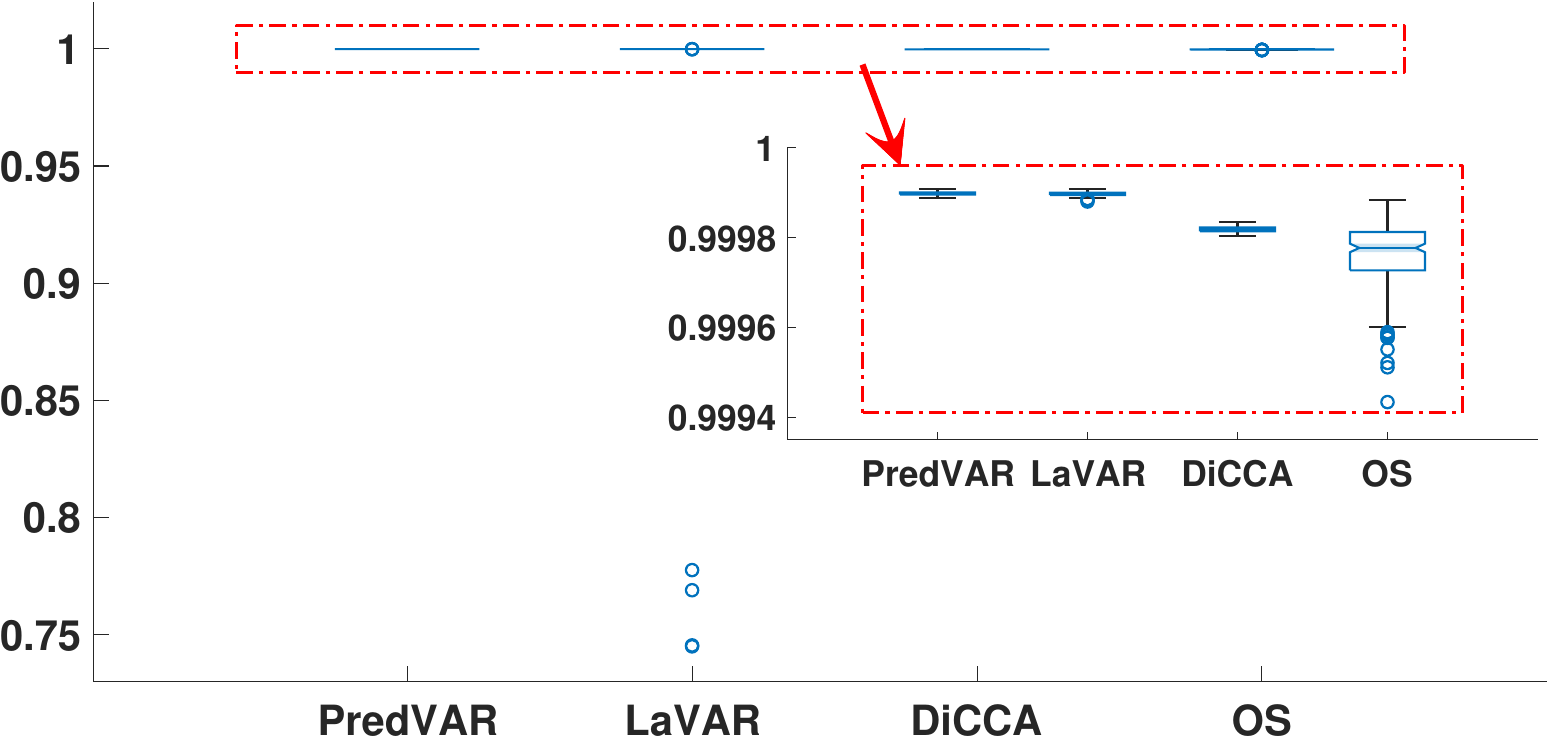}
         \caption{Average correlation between~$\{\mathbf{\hat P}\bm{ \hat v}_{k|\mathbf{\hat R}}\}$ and~$\{\mathbf{\hat P}\bm{ v}_{k|\mathbf{\hat R}}\}$.}
         \label{fig:Rep_in}
     \end{subfigure}
        \caption{Performance of \modelname, LaVAR-CCA, DiCCA, and OS over multiple measurement time series.}
        \label{fig:Rep}
\end{figure*}
\begin{figure}
    \centering
     \includegraphics[width=0.82\columnwidth]{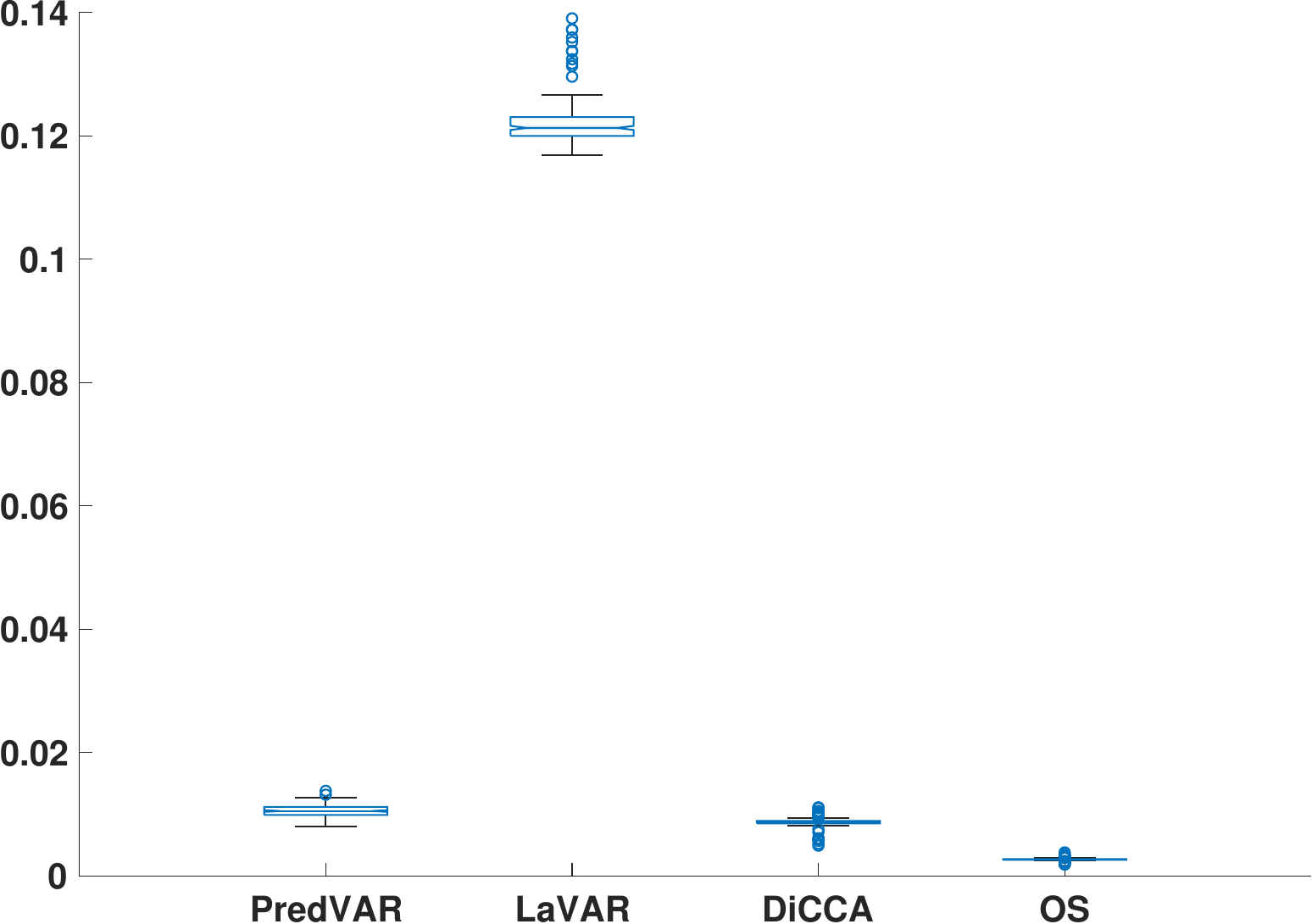}
     \caption{Training time of various algorithms }
     \label{fig:train_time}
\end{figure}


\begin{table*}[t]
\caption{D-distance between the ranges of~$\mathbf{\hat P}$ estimated by PredVAR, LaVAR-CCA, and DiCCA as~$s$ or~$\ell$ varies.}
    \label{tab:Eastman_order}
    \scriptsize
    \setlength\tabcolsep{5pt}
        \renewcommand{\arraystretch}{1.1}
            \begin{tabularx}{0.97\linewidth}{c|cccccccccccc}
  \toprule    \toprule
$s $ (with $\ell=14$) & 1&2&3&4&5&6&7&8&9&10&11&12 \\\hline
PredVAR~--~LaVAR & 0.0001    &0.0001  &  0.0000   & 0.0000   & 0.0001  &  0.0002   & 0.0000  &  0.0000    &0.0000  &  0.0000  &  0.0002  &  0.0000 \\
PredVAR~--~DiCCA & 0.0539    &0.0297  &  0.0480   & 0.0320   & 0.0308  &  0.0392   & 0.0440  &  0.0378    &0.0401  &  0.0486  &  0.0515  &  0.0518 \\
LaVAR~--~DiCCA & 0.0539    &0.0297  &  0.0480   & 0.0320   & 0.0308  &  0.0393   & 0.0440  &  0.0378    &0.0401  &  0.0486  &  0.0514  &  0.0518\\
  \bottomrule \bottomrule

 $\ell $ (with $s=5$)  & 6&7&8&9&10&11&12&13&14&15&16&17 \\\hline
PredVAR~--~LaVAR & 0.1494    &0.1203  &  0.3430   & 0.3234   & 0.0078  &  0.0004   & 0.0005  &  0.0008    &0.0000  &  0.0008  &  0.0002  &  0.0045 \\
PredVAR~--~DiCCA & 0.4287    &0.4055  &  0.2977   & 0.1990   & 0.2037  &  0.0961   & 0.1283  &  0.0538    &0.0518  &  0.1146  &  0.2427  &  0.1350 \\
LaVAR~--~DiCCA & 0.4104    &0.3949  &  0.2284   & 0.3392   & 0.2009  &  0.0960   & 0.1285  &  0.0541    &0.0518  &  0.1151  &  0.2427  &  0.1314\\
  \bottomrule \bottomrule
  \end{tabularx}
\end{table*}
\subsubsection{The DLD~$\ell$ and the VAR order~$s$} Given the estimated model sizes~$(\ell,s)$, we first identify the \modelname~parameters via the proposed algorithm using the training data and then validate the performance of the estimated model using the testing data.
To compare between two loadings matrices
$\mathbf{P}_1 \in \Re^{p\times l_1}$ and~$\mathbf{P}_2 \in \Re^{p\times l_2}$, the \emph{D-distance} between them defined in \cite{pan:yao:2008,gao:tsay:2021high-Dim} is adopted and extended for the case of $l_1 \geq l_2$ as follows
\[
\sqrt{1-\textrm{trace}\left(\mathbf{P}_1(\mathbf{P}_1^\intercal\mathbf{P}_1)^{-1}\mathbf{P}_1 ^\intercal\mathbf{P}_2(\mathbf{P}_2^\intercal\mathbf{P}_2)^{-1}\mathbf{P}_2 ^\intercal\right)/l_1}.
\]
We further
define the average correlation between $\{\bm y_k\}$ and~$\{\bm y'_k\}$ by the average absolute value of the pair-wise correlations, i.e.,
\[
\sum\nolimits_{i=1}^p |\text{corr}(y_i,y'_i)|/p.
\]
With the two measures, Table~\ref{tab:DlDs} shows the performances of our approach under the different~$(\ell,s)$-pairs. Since $\bm {v}_k$ depends on the latent coordinates from various methods, we map it to the original data space by the loadings matrix~$\mathbf{P}$ or its estimate~$\mathbf{\hat P}$ to compare various models.

It is easily seen from Tables~\ref{tab:DlDs_angleP}~--~\ref{tab:DlDs_in} that~$\ell=3$, which is also the truth, gives the highest average correlation between the original noise-free signal series~$\{\mathbf{P}\bm {v}_k\}$ and the signal reconstruction series~$\{\mathbf{\hat P}\bm{ v}_{k|\mathbf{\hat R}}\}$ (Table~\ref{tab:DlDs_proj}) or the signal prediction series~$\{\mathbf{\hat P}\bm{\hat v}_{k|\mathbf{\hat R}}\}$ (Table~\ref{tab:DlDs_pred}).

Since the noise-free  series~$\{\mathbf{P}\bm {v}_k\}$ is unavailable in practical scenarios, Table~\ref{tab:DlDs_in}~--~\ref{tab:DlDs_inout} deserves particular attention.  Specifically, for each~$s$, the average correlation between~$\{\mathbf{\hat P}\bm{\hat v}_{k|\mathbf{\hat R}}\}$ and~$\{\mathbf{\hat P}\bm{v}_{k|\mathbf{\hat R}}\}$ significantly deceases as the overestimate of~$\ell$ occurs. Moreover, for a proper~$s$, the average correlation between~$\{\mathbf{\hat P}\bm{\hat v}_{k|\mathbf{\hat R}}\}$ and~$\{\bm y_k\}$  achieves the maximum value when~$\ell=3$.


\subsubsection{Comparison to other Methods}
Using the same DLV series and loadings matrices, we perform a Monte-Carlo simulation of the noise from the same distribution to obtain~$200$ series of~$10,000$ points each. For each time series, the first~$7,000$ data points are for training and the rest for testing. Uniformly set~$\ell =3$ and~$s=2$. Fig.~\ref{fig:Rep} shows the performance of the PredVAR and the benchmark algorithms on the testing data series.

Figs.~\ref{fig:Rep_angleP} and~\ref{fig:Rep_proj} show that the \modelname, LaVAR, and DiCCA algorithms are comparable in identifying the dynamic subspace and reconstructing the uncorrupted signal. In contrast, the OS algorithm has notably inferior performance. Moreover, the OS algorithm exhibits more significant variability regarding subspace identification and signal reconstruction than the other algorithms.

Fig.~\ref{fig:Rep_pred} showcases the performance of various algorithms in predicting the noise-free signal, with the \modelname~approach performing the best. In most cases, the LaVAR algorithm performs similarly to \modelname~and surpasses DiCCA and OS. However, LaVAR exhibits more significant outliers than the other three algorithms, meaning that LaVAR is numerically less reliable. A similar phenomenon is observed in Fig.~\ref{fig:Rep_in}, regarding the average correlation between the reconstruction and prediction time series. Furthermore, in Fig.~\ref{fig:Rep_in}, DiCCA and OS perform significantly worse than \modelname~and LaVAR in general. This fact strongly verifies the importance of exploring the interactions among different DLVs and the alternating updates of projection-related and dynamics-related parameters.



We also compare the algorithms in terms of training time. As shown in Fig.~\ref{fig:train_time}, OS requires the least training effort, and LaVAR demands the most. One would expect DiCCA to have less training time than \modelname~and LaVAR since DiCCA does not involve interactions among DLVs.  Surprisingly, Fig.~\ref{fig:train_time} suggests that  \modelname~takes slightly less training time than DiCCA. However, a unique advantage of DiCCA over other interacting DLV models is that, when a DiCCA model with $\ell$ DLVs is obtained, it yields a suite of DiCCA models for $1:\ell$. For algorithms with interacting DLV models, one has to build one model with each DLD from 1 to  $\ell$.


\subsection{Case Study on Industrial Process Data}
In this study, the plant-wide oscillation dataset from the Eastman Chemical Company~\cite{Qin:2022LaVAR_AIChEJ} is used to demonstrate the effectiveness of the proposed~\modelname~algorithm, where~$18$ variables with oscillation patterns are selected for modeling over their first $1,000$ sequential samples.

The RRMFPE criterion suggests~$\ell=14$ and~$s=5$. With such model sizes, the DLVs extracted by the PredVAR algorithm are plotted in Fig.~\ref{fig:Eastman}. The first three DLVs prominently display low-frequency oscillations. The last DLV seems to exhibit the highest volatility.
Moreover, Table~\ref{tab:Eastman_order} depicts the D-distances between the dynamic subspaces estimated by PredVAR, LaVAR-CCA, and DiCCA as~$s$ or~$\ell$ varies. Similarly to before, the difference is more responsive to the change of~$\ell$ than~$s$. The dynamic subspaces obtained by PredVAR and LaVAR-CCA are more similar than those estimated by DiCCA.

\begin{figure}[t]
    \centering
\includegraphics[width=1.02\columnwidth]{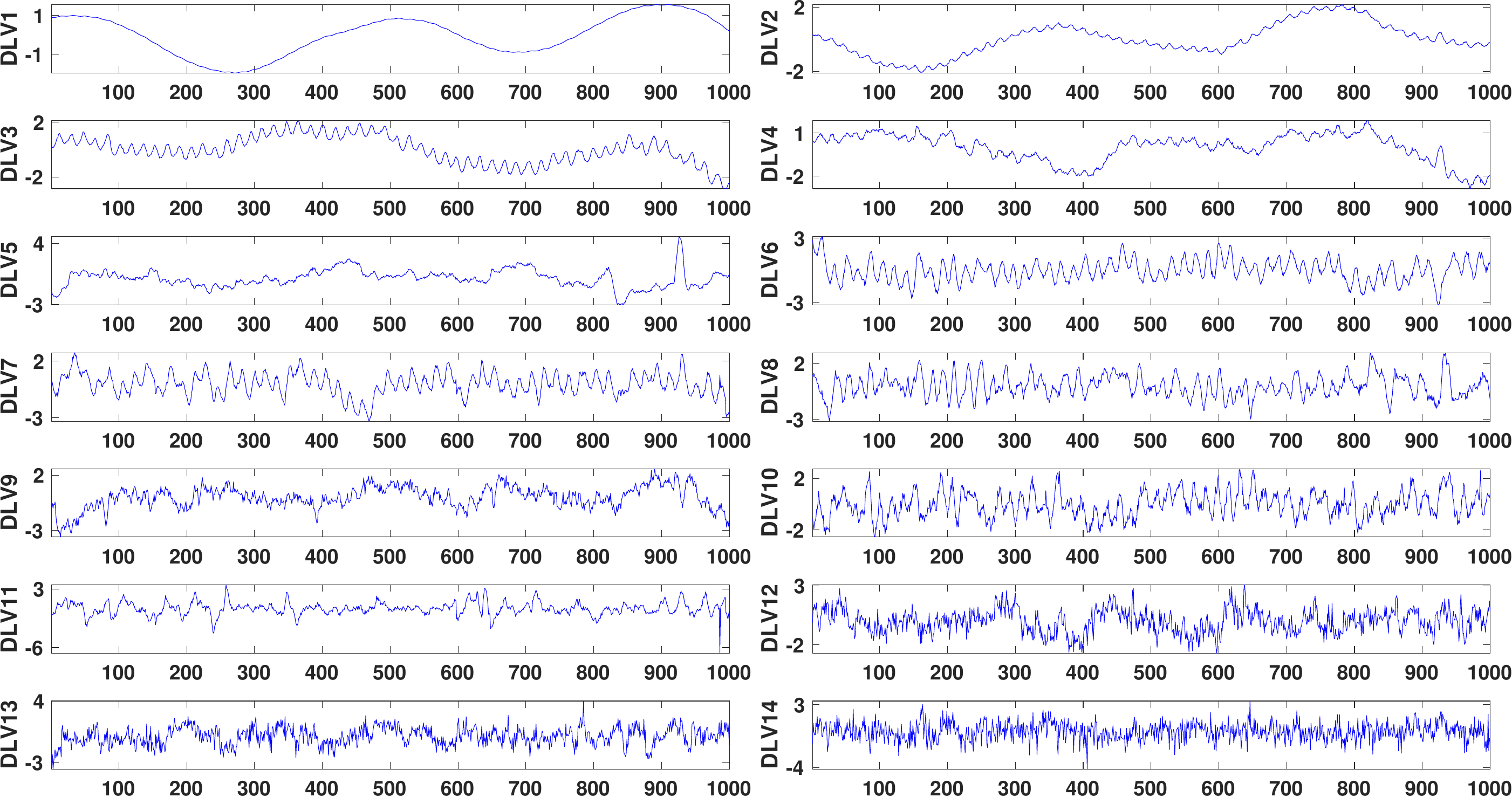}
    \caption{DLVs extracted by the PredVAR algorithm with~$\ell =14$ and $s=5$ for the Eastman Process Data.}
    \label{fig:Eastman}
\end{figure}




\section{Conclusions} \label{sec:conclusion}

A probabilistic reduced dimensional VAR modeling algorithm, namely, \modelname, is successfully developed with oblique projections, which is required for optimal low-dimensional dynamic modeling from high-dimensional data. The \modelname~model is equivalent to a reduced-rank VAR model, where the VAR coefficient matrices have reduced rank. Both model forms admit a serially independent subspace of the measurement vector.
The algorithm is iterative, following the principle of expectation maximization, which alternately estimates the DLV dynamic model and the outer oblique projection. We show with a simulated case and an industrial case study that the  \modelname~algorithm  with the oblique projections outperforms other recently developed algorithms, including a non-iterative one for reduced-dimensional dynamic modeling. The computational cost of the EM-based PredVAR algorithm is much reduced compared to the alternating optimization based LaVAR algorithm in \cite{Qin:2022LaVAR_AIChEJ}.

\bibliographystyle{abbrv}


\appendix
\section{Proof of Theorem~\ref{thm:eq_two_models}} \label{apd:eq_two_models}

Proof of Part 1). From the \modelname~model \eqref{eq:outer_model} and \eqref{eq:DLV_VAR} it is straightforward to have
\begin{align*}
  \bm y_k &=\sum_{j=1}^s\mathbf{ P}\mathbf{ B}_j \bm v_{k-j} +\mathbf{ P} \bm \varepsilon_k + \mathbf{\bar P} \bm{\bar \varepsilon}_k \\
  &=\sum_{j=1}^s\mathbf{ P}\mathbf{ B}_j\mathbf{ R}^\intercal \bm y_{k-j} +\bm e_k
\end{align*}
with $\mathbf{ R}^\intercal\mathbf{ P} = \mathbf{ I}$, where  $\bm e_k = \mathbf{ P} \bm \varepsilon_k + \mathbf{\bar P} \bm{\bar \varepsilon}_k$. Since $ \bm \varepsilon_k$ and $\bm{\bar \varepsilon}_k$ are i.i.d. Gaussian, $\bm e_k\sim \mathcal{N}(\bm 0, \mathbf{\Sigma}_{\bm{e}})$ is i.i.d. Gaussian.

From the canonical \rrname~model, pre-multiplying~$\mathbf{R}^\intercal$ to~\eqref{eq:can_RRVAR} leads to
\begin{align*}
 \mathbf{R}^\intercal \bm y_k &=\sum_{j=1}^s\mathbf{R}^\intercal\mathbf{ P}\mathbf{ B}_j \mathbf{R}^\intercal\bm y_{k-j} +\mathbf{R}^\intercal \bm{e}_k \\
  \bm{v}_k &=\sum_{j=1}^s\mathbf{ P}\mathbf{ B}_j \bm v_{k-j} +\bm{\varepsilon}_k,
\end{align*}
where~$\bm{v}_k=\mathbf{R}^\intercal\bm{y}_k$ and~$\bm{\varepsilon}_k=\mathbf{R}^\intercal \bm{e}_k$. Since $\bm{e}_k$ is i.i.d. Gaussian, $\bm \varepsilon_k\sim \mathcal{N}(\bm 0, \mathbf{R}^\intercal \mathbf{\Sigma}_{\bm{e}}\mathbf{R})$ is i.i.d. Gaussian.

Let~$\mathbf{\bar R}\in \Re^{p\times(p-\ell)}$ be an arbitrary matrix that is of full column rank and~$\mathbf{\bar R}^\intercal \mathbf{P}=\bm 0$. It follows that~$\bm{\bar \varepsilon}_k= \mathbf{\bar R}^\intercal \bm y_k= \mathbf{\bar R}^\intercal \bm e_k$. Since $\bm{e}_k$ is i.i.d. Gaussian, $\bm{\bar \varepsilon}_k\sim \mathcal{N}(\bm 0, \mathbf{\bar R}^\intercal \mathbf{\Sigma}_{\bm{e}}\mathbf{\bar R})$ is i.i.d. Gaussian. By~\eqref{eq:RP_relation}, obtain~$\mathbf{\bar P}$ by
\[
\mathbf{\bar P} = \begin{bmatrix}
    \mathbf{R}^\intercal \\ \mathbf{\bar R}^\intercal
\end{bmatrix}^{-1}\begin{bmatrix}\bm 0\\ \mathbf{I}\end{bmatrix},
\]
and it holds that
\[
\bm{y}_k = \mathbf{P}\mathbf{R}^\intercal \bm{y}_k+\mathbf{\bar P}\mathbf{\bar R}^\intercal \bm{y}_k = \mathbf{P}\bm{v}_k +\mathbf{\bar P} \bm{\bar \varepsilon}_k.
\]
Thus, the canonical \rrname~model is equivalently transformed into the \modelname~model \eqref{eq:outer_model} and \eqref{eq:DLV_VAR}.

Proof of Part 2). From \eqref{eq:RRVAR} we have
\begin{align*}
\bm y_k &=\sum_{j=1}^s\mathbf{\acute P}\mathbf{\acute B}_j\mathbf{\acute R}^\intercal \bm y_{k-j} +\bm e_k \\
 &=\sum_{j=1}^s\mathbf{\acute P}\mathbf{\acute B}_j
 (\mathbf{\acute R}^\intercal\mathbf{\acute P})
 (\mathbf{\acute R}^\intercal\mathbf{\acute P})^{-1} \mathbf{\acute R}^\intercal \bm y_{k-j} +\bm e_k \\
 &=\sum_{j=1}^s\mathbf{ P}\mathbf{ B}_j\mathbf{ R}^\intercal \bm y_{k-j} +\bm e_k
\end{align*}
where $\mathbf{ R}^\intercal\mathbf{ P} = \mathbf{ I}$ by setting
   $\mathbf{P} =\mathbf{\acute P} $,
   $ \mathbf{B}_j =\mathbf{\acute B}_j \mathbf{\acute R}^\intercal\mathbf{\acute P}$,
   and
$ \mathbf{R} =\mathbf{\acute R}(\mathbf{\acute P}^\intercal\mathbf{\acute R})^{-1}$. Further,
\[
\mathbf{R}^\intercal\mathbf{P}= (\mathbf{\acute R}^\intercal\mathbf{\acute P})^{-1} \mathbf{\acute R}^\intercal \mathbf{\acute P}=\mathbf{I}.
\]


\section{Proof of Theorem~\ref{thm:eq_constraint}} \label{apd:eq_constraint}

Applying the Lagrangian multiplier~$\mathbf{\Gamma}$ to~\eqref{eq:Sigma_varepsilon} leads to
\[
L(\mathbf{R},\mathbf{\Gamma}) = {\mathbf{R}}^\intercal\mathbf{\Sigma}_{\bm e}\mathbf{R}+\mathbf{\Gamma}(\mathbf{I} - {\mathbf{R}}^\intercal\mathbf{P})
\]
Differentiating~$L$ with respect to~$\mathbf{R}$ and setting the derivative to zero gives
\begin{equation}\label{eq:diff_R}
    \frac{\partial L}{\partial \mathbf{R}}= 2\mathbf{\Sigma}_{\bm e}\mathbf{R} - \mathbf{P}\mathbf{\Gamma}=\bm 0.
\end{equation}
Since~${\mathbf{R}}^\intercal\mathbf{P}=\mathbf{I}$, pre-multiplying~${\mathbf{R}}^\intercal$ to~\eqref{eq:diff_R} gives
\begin{equation}\label{eq:Sigma}
    \mathbf{\Gamma}= 2{\mathbf{R}}^\intercal\mathbf{\Sigma}_{\bm e}\mathbf{R}.
\end{equation}
Substituting~\eqref{eq:Sigma} into~\eqref{eq:diff_R} leads to 1). It follows from the Lagrangian multiplier analysis~\cite{boyd2004convex} that Problem~\eqref{eq:Sigma_varepsilon} attains the optimum only if 1) holds. It remains to show the equivalence of the six statements.

 1) $\Leftrightarrow$~2) is due to that~$\mathbf{P}$ is of full column rank and post-multiplying~$\mathbf{P}^\intercal$ to~1) leads to
    \[
     (\mathbf{I}-\mathbf{P}\mathbf{R}^\intercal)\mathbf{\Sigma}_{\bm e}\mathbf{R}\mathbf{P}^\intercal
         = E \left ( (\mathbf{I} - \mathbf{P}{\mathbf{R}}^\intercal){\bm e}_k {{\bm e}_k}^\intercal \mathbf{R}{\mathbf{P}}^\intercal \right ) =\bm 0.
    \]
    2) $\Leftrightarrow$~3) follows from~\eqref{eq:RP_relation}, namely
    \begin{align*}
        &(\mathbf{I}-\mathbf{P}\mathbf{R}^\intercal)\bm{e}_k=\mathbf{\bar P}\mathbf{\bar R}^\intercal\bm{e}_k=\mathbf{\bar P}\bm{\bar \varepsilon}_k\\
        &\mathbf{\bar R}^\intercal((\mathbf{I}-\mathbf{P}\mathbf{R}^\intercal)\bm{e}_k)=\mathbf{\bar R}^\intercal\bm{e}_k =\bm{\bar \varepsilon}_k\\
        &\mathbf{R}^\intercal(\mathbf{P}\mathbf{R}^\intercal\bm{e}_k) = \mathbf{R}^\intercal\bm{e}_k=\bm{\varepsilon}_k
    \end{align*}
    3) $\Leftrightarrow$~4) follows from the definitions of $\bm{\varepsilon}_k$, $\bm{\bar \varepsilon}_k$, and~$\mathbf{\Sigma}_{\bm e}$.

    4) $\Leftrightarrow$~5) follows from the fact that the inverse of a block diagonal matrix is also block diagonal, and~\eqref{eq:Sigmae_inverse}.

    4) $\Leftrightarrow$~6) follows from~$\mathbf{P}^\intercal \mathbf{\bar R}=\bm 0$ and
    \begin{equation*}
        \mathbf{\Sigma}_{\bm y}=\mathbf{\Sigma}_{\bm e} + \mathbf{P}\mathbf{\Sigma}_{\bm{\hat v}}\mathbf{P}^\intercal.
    \end{equation*}
6) $\Leftrightarrow$~7) follows from the equivalence among
    \begin{align*} \label{eq:Sigma_y}
\left[
      \mathbf{R}~~ \mathbf{\bar{R}}
\right]^\intercal\mathbf{\Sigma}_{\bm y} \left[
      \mathbf{R}~~ \mathbf{\bar{R}}
\right] =
\begin{bmatrix}
    \mathbf{\Sigma}_{\bm v} & \bm 0\\
    \bm 0 & \mathbf{\Sigma}_{\bm{\bar \varepsilon}}
\end{bmatrix},
 \end{align*}
\begin{equation*} \label{eq:Sigmay_inverse}
\mathbf{\Sigma}_{\bm y}^{-1} =
\left[
      \mathbf{R}~~ \mathbf{\bar{R}}
\right]
\begin{bmatrix}  \mathbf{\Sigma}_{\bm v}^{-1} \\
~ & \mathbf{\Sigma}_{\bm{\bar \varepsilon}}^{-1}
    \end{bmatrix}
\left[
      \mathbf{R}~~ \mathbf{\bar{R}}
\right]^{\intercal},
\end{equation*}
 if $\mathbf{\Sigma}_{\bm y}$ is non-singular, and
\begin{equation*}
\left[
      \mathbf{P}~~ \mathbf{\bar{P}}
\right]^\intercal \mathbf{\Sigma}_{\bm y}^{-1} \left[
      \mathbf{P}~~ \mathbf{\bar{P}}
\right]=
\begin{bmatrix}  \mathbf{\Sigma}_{\bm v}^{-1} \\
~ & \mathbf{\Sigma}_{\bm{\bar \varepsilon}}^{-1}
    \end{bmatrix}.
\end{equation*}
\end{document}